\documentclass[12pt,english]{elsarticle}
\usepackage[T1]{fontenc}
\usepackage[a4paper]{geometry}
\usepackage{color}
\usepackage{array}
\usepackage{float}
\usepackage{booktabs}
\usepackage{textcomp}
\usepackage{multirow}
\usepackage{amsmath}
\usepackage{amssymb}
\usepackage{graphicx}
\usepackage{stmaryrd}
\usepackage{mfirstuc}
\usepackage{lineno}   % for numberline
\usepackage{setspace} % for spacing
\usepackage[colorlinks,allcolors=blue]{hyperref}
\usepackage{breqn}
\journal{arXiv}
\newcommand{\Phase}[1]{\Statex\hspace*{-\algorithmicindent}\textbf{#1}}

\makeatletter

\providecommand{\tabularnewline}{\\}
\floatstyle{ruled}
\newfloat{algorithm}{tbp}{loa}
\providecommand{\algorithmname}{Algorithm}
\floatname{algorithm}{\protect\algorithmname}

\usepackage{algorithm}
\usepackage{algpseudocode}

\providecommand{\tabularnewline}{\\}

\@ifundefined{showcaptionsetup}{}{%
	\PassOptionsToPackage{caption=false}{subfig}}
\usepackage{subfig}
\makeatother

\usepackage{babel}
\begin{document}
\doublespacing
\begin{frontmatter}{}

\title{Data-Guided Physics-Informed Neural Networks for Solving Inverse
Problems in Partial Differential Equations}

\author[UC]{Wei~Zhou \corref{cor1}}
\ead{zhouw6@mail.uc.edu}
\author[UC]{Y.F.~Xu}
\ead{yongfengxuyf@gmail.com}
\address[UC]{Department of Mechanical and Materials Engineering, University of Cincinnati, Cincinnati, OH 45221, USA}
\cortext[cor1]{Corresponding author}

\begin{abstract}
Physics-informed neural networks (PINNs) represent a significant advancement
in scientific machine learning by integrating fundamental physical
laws into their architecture through loss functions. PINNs have
been successfully applied to solve various forward and inverse problems
in partial differential equations (PDEs). However, a notable challenge can
emerge during the early training stages when solving inverse problems.
Specifically, data losses remain high while PDE residual losses are minimized rapidly, thereby exacerbating the imbalance between loss terms and impeding the overall efficiency
of PINNs. To address this challenge, this study proposes a novel framework
termed data-guided physics-informed neural networks (DG-PINNs).
The DG-PINNs framework is structured into two distinct phases: a pre-training phase and a fine-tuning
phase. In the pre-training phase, a loss function with only the
data loss is minimized in a neural network. In the fine-tuning phase,
 a composite loss function, which consists of the data loss, PDE
residual loss, and, if available, initial and boundary condition losses,
is minimized in the same neural network. Notably, the pre-training phase ensures that the data loss is
already at a low value before the fine-tuning phase commences. This approach enables the fine-tuning phase to converge to a minimal
composite loss function with fewer iterations compared to existing
PINNs. To validate the effectiveness, noise-robustness, and efficiency
of DG-PINNs, extensive numerical investigations are conducted on inverse
problems related to several classical PDEs, including the heat equation,
wave equation, Euler--Bernoulli beam equation, and Navier--Stokes
equation. The numerical results demonstrate that DG-PINNs can accurately solve these inverse problems and exhibit robustness against noise in training data. Furthermore, it is shown that DG-PINNs can significantly enhance
the efficiency in solving inverse problems compared to existing
PINNs while maintaining high solution accuracy. For transparency
and reproducibility, all codes and datasets used in this study are
freely available at \url{https://github.com/dopawei/DG-PINNs}.
\end{abstract}

\end{frontmatter}{}

\section{Introduction\label{sec:1}}

In recent years, physics-informed machine learning \citep{karniadakis2021physics},
particularly physics-informed neural networks (PINNs) \citep{raissi2019physics},
has become a promising method for solving complex problems involving
partial differential equations (PDEs). PINNs leverage the universal approximation capabilities of neural networks \citep{hornik1989multilayer}
and automatic differentiation \citep{baydin2018automatic} to approximate
the latent solutions of PDEs. The fundamental physical laws are integrated
into loss functions of neural networks through automatic differentiation
\citep{raissi2019physics}. This integration can be traced back to
the 1990s \citep{lee1990neural,dissanayake1994neural,meade1994solution},
which enables neural network outputs to comply with fundamental physical
laws. PINNs have been applied to a diverse range of fields, including
fluid dynamics \citep{raissi2020hidden,mao2020physics,jin2021nsfnets,fang2022immersed},
continuum mechanics \citep{shukla2020physics,haghighat2021physics,zhou2023damageIMAC,zhou2024damage},
biomedical engineering \citep{kissas2020machine}, and others \citep{fang2019deep,chen2020physics,ji2021stiff,guo2023structural}.

Despite the broad applicability of PINNs, several challenges remain
in training PINNs efficiently and effectively. Issues such as imbalanced
gradient magnitudes of loss terms \citep{wang2021understanding} and
inherent spectral biases \citep{rahaman2019spectral,xu2019frequency}
within PINNs can impede convergence and result in poor solutions \citep{wang2021eigenvector,wang2022and}.
Recent efforts have focused on enhancing the trainability of PINNs.
Specifically, for the challenge of imbalanced gradient magnitudes,
Wang et al. \citep{wang2021understanding} have identified this as
a significant factor in the failure of PINNs. They proposed a learning
rate annealing algorithm as a solution to balance gradient magnitudes
of loss terms. Besides, Wang et al. \citep{wang2022and} explored
the training dynamics of PINNs through the lens of the neural tangent
kernel (NTK) theory \citep{jacot2018neural} and found a notable discrepancy
in the convergence rate of each loss term. They introduced a novel
adaptive weights algorithm that utilizes the eigenvalues of the NTK
to adaptively adjust loss weights. Furthermore, Li et al. \citep{li2023physics}
observed that not only gradient magnitudes of loss terms are often imbalanced,
but also the gradient directions of loss terms are usually conflicting.
To address both issues, they introduced an adaptive gradient descent method for balancing gradient magnitudes and eliminating conflicts in gradient directions. For the challenge of inherent spectral
biases in PINNs, approaches such as Fourier feature networks \citep{tancik2020fourier}
have been applied for PDEs involving high-frequency components in
their latent solutions \citep{wang2021eigenvector,song2023simulating}.
Additionally, several methods have been proposed to enhance the training
efficiency and performance of PINNs, including sinusoidal input mapping
\citep{wong2022learning}, residual-based adaptive refinement \citep{lu2021deepxde},
the coupled-automatic-numerical differentiation framework \citep{chiu2022can},
and other variants of PINNs \citep{jagtap2020adaptive,dwivedi2020physics,basir2022physics,tseng2023cusp,mcclenny2023self}.

In solving inverse problems using existing PINNs, the primary objective
is to accurately estimate unknown parameters within PDEs from observed
data. To achieve this objective, PINNs are trained to minimize a composite
loss function, which includes the PDE residual loss, initial condition
loss, boundary condition loss and data loss \citep{raissi2019physics,mao2020physics,shukla2020physics,lu2021physics,zhang2022analyses,rasht2022physics,xu2023transfer,zhou2023damageIMAC}.
For instance, Mao et al. \citep{mao2020physics} employed PINNs to
estimate density, velocity, and pressure fields for one-dimensional
Euler equations from observed density gradient data. Besides, Rasht
et al. \citep{rasht2022physics} applied PINNs for full waveform inversions
in seismic imaging to estimate the wave speed from observed data.
It is noted that the data loss is derived from neural network outputs,
while the PDE residual loss is based on both derivatives of these
neural network outputs and unknown parameters within PDEs. Besides,
initial neural network outputs in PINNs tend to approximate flat output
functions, which can be governed by numerous PDEs \citep{wong2022learning}.
Hence, during the early stages of training PINNs, the PDE residual loss
can be minimized to a negligible level, yet the data loss remains at a high
level. This imbalance between the loss terms requires additional training
iterations to both reduce the data loss and accurately estimate unknown
parameters, thus impeding the efficiency of solving inverse problems
by PINNs.

To address the challenge of solving inverse problems with PINNs, this
paper introduces a novel framework termed data-guided physics-informed neural networks (DG-PINNs). The
DG-PINNs framework consists of two phases. The first phase is the
pre-training phase: a neural network is trained to minimize the data loss. The second phase is the fine-tuning phase: the neural network
is further optimized to simultaneously minimize the composite loss
function. To assess the effectiveness, noise-robustness, and efficiency
of DG-PINNs, extensive numerical investigations are conducted on various
PDEs, including the heat equation, wave equation, Euler--Bernoulli
beam equation, and Navier--Stokes equation. For each equation, sensitivity
analyses of the DG-PINNs results are conducted, focusing on two hyperparameters: the maximum
number of iterations in the pre-training phase and the number of data
points in training datasets for the data loss. Besides, a comparative
analysis between DG-PINNs and PINNs is conducted for each investigated
equation.

The paper is organized as follows: Sections 2.1 through 2.3 provide
a brief overview of the fundamentals of fully-connected neural networks,
PINNs for inverse problems, and the adaptive weights algorithm, respectively.
Section 2.4 details the proposed DG-PINNs. Section 3 presents extensive
numerical investigations, illustrating the effectiveness, noise-robustness
and efficiency of DG-PINNs in various inverse problems. Finally, Section
4 offers concluding remarks and future research directions.

\section{Methodology\label{sec:2}}

\subsection{Fully-connected neural networks}

Consider a $H$-layer fully connected feed-forward neural network,
which consists of $H$ hidden layers. This neural network can be expressed
in a recurrent form as follows:

\begin{equation}
\mathbf{f}^{\left(h\right)}=\sigma\left(\mathbf{W}^{\left(h\right)}\mathbf{f}^{\left(h-1\right)}+\mathbf{b}^{\left(h\right)}\right)\text{,}\quad h=1,2,...,H,\label{eq:NN}
\end{equation}
where $\mathbf{f}^{\left(h\right)}$ denotes the output of the $h$-th
hidden layer, $\mathbf{f}^{\left(0\right)}$ is the neural network
input, $\sigma\left(\cdot\right)$ is the nonlinear activation function,
$\mathbf{W}^{\left(h\right)}$ and $\mathbf{b}^{\left(h\right)}$
denote the weight matrix and bias vector for the $h$-th hidden layer,
respectively. The neural network output is given by

\begin{equation}
\hat{\mathbf{u}}=\mathbf{W}^{\left(H+1\right)}\mathbf{f}^{\left(H\right)}+\mathbf{b}^{\left(H+1\right)},
\end{equation}
where $\mathbf{W}^{\left(H+1\right)}$ and $\mathbf{b}^{\left(H+1\right)}$
are the weight matrix and bias vector of the output layer, respectively.
The set of all parameters, including weights and biases within the
network, can be grouped and denoted by

\begin{equation}
\boldsymbol{\theta}=\left\{ \mathbf{\mathbf{W}}^{\left(1\right)},\mathbf{b}^{\left(1\right)},...,\mathbf{W}^{\left(h\right)},\mathbf{b}^{\left(h\right)},...,\mathbf{\mathbf{W}}^{\left(H+1\right)},\mathbf{b}^{\left(H+1\right)}\right\} .\label{eq:parameters}
\end{equation}
The architecture of this neural network integrates linear operations
with nonlinear activation functions. When the activation functions
are infinitely differentiable, such as sine, cosine, and hyperbolic
functions, this architecture allows the computation of meaningful
derivatives of any order for the output $\hat{\mathbf{u}}$ with respect
to the neural network input through the automatic differentiation \citep{baydin2018automatic}. Such derivatives facilitate the integration
of fundamental physical laws into loss functions in PINNs.

\subsection{PINNs for solving inverse problems}

PINNs offer a novel method for solving inverse problems in PDEs of
various systems. Such a typical inverse problem is formulated as
\begin{equation}
\mathcal{F}\left[u\left(\boldsymbol{x},t\right);\boldsymbol{\gamma}\right]=s\left(\boldsymbol{x},t\right),\quad\boldsymbol{x}\in\Omega,\:t\in\left[0,T\right],\label{eq:PDEs}
\end{equation}
where $\mathcal{\mathcal{F}}\left[\cdot\right]$ is the differential
operator, $u\left(\boldsymbol{x},t\right)$ denotes the observed data,
$\boldsymbol{x}=\left[x_{1},...,x_{n}\right]$ is the spatial coordinates,
$t$ is the time variable over the duration $T$, $\boldsymbol{\gamma}$
are parameters related to the physics of the system, all or some of
which are unknown parameters for inverse problems, and $s\left(\boldsymbol{x},t\right)$
is the source function. The initial conditions of Eq. (\ref{eq:PDEs})
can be expressed by

\begin{equation}
\mathcal{I}\left[u\left(\boldsymbol{x},t\right)\right]=f\left(\boldsymbol{x}\right),\quad\boldsymbol{x}\in\Omega,\text{\;}t=0,\label{eq:ICs}
\end{equation}
and the boundary condition of Eq. (\ref{eq:PDEs}) can be expressed
by

\begin{equation}
\mathcal{B}\left[u\left(\boldsymbol{x},t\right)\right]=q\left(\boldsymbol{x},t\right),\quad\boldsymbol{x}\in\partial\Omega,\text{\;}t\in\left[0,T\right],\label{eq:BCs}
\end{equation}
where $\mathcal{I}\left[\cdot\right]$ and $f\left(\boldsymbol{x}\right)$
denote the initial condition operator and initial condition function,
respectively, and $\mathcal{B}\left[\cdot\right]$ and $q\left(\boldsymbol{x},t\right)$
denote the boundary condition operator and boundary condition function,
respectively.

For solving inverse problems using PINNs, the aim is to accurately
estimate $\boldsymbol{\gamma}$. To achieve this aim, PINNs are trained
to minimize a composite loss function \citep{raissi2019physics},
which is expressed by

\begin{equation}
\mathcal{L}\left(\boldsymbol{\Theta}\right)=\lambda_{r}\mathcal{L}_{r}\left(\boldsymbol{\Theta}\right)+\lambda_{i}\mathcal{L}_{i}\left(\boldsymbol{\theta}\right)+\lambda_{b}\mathcal{L}_{b}\left(\boldsymbol{\theta}\right)+\lambda_{d}\mathcal{L}_{d}\left(\boldsymbol{\theta}\right),\label{eq:loss_function}
\end{equation}
where $\boldsymbol{\Theta}=\left\{ \boldsymbol{\theta},\boldsymbol{\gamma}\right\} $
with the total number of parameters $M$, $\lambda_{r}$, $\lambda_{i}$,
$\lambda_{b}$ and $\lambda_{d}$ are loss weights corresponding to
the PDE residual loss $\mathcal{L}_{r}\left(\boldsymbol{\Theta}\right)$,
initial condition loss $\mathcal{L}_{i}\left(\boldsymbol{\theta}\right)$,
boundary condition loss $\mathcal{L}_{b}\left(\boldsymbol{\theta}\right)$,
and data loss $\mathcal{L}_{d}\left(\boldsymbol{\theta}\right)$,
respectively. With the prediction of PINNs, $\hat{u}\left(\boldsymbol{x},t;\boldsymbol{\theta}\right)$,
the aforementioned individual loss terms are expressed by

\begin{equation}
\mathcal{L}_{r}\left(\boldsymbol{\Theta}\right)=\frac{1}{N_{r}}\sum_{k=1}^{N_{r}}\left(\mathcal{F}\left[\hat{u}\left(\boldsymbol{x}_{r}^{k},t_{r}^{k};\boldsymbol{\theta}\right);\boldsymbol{\gamma}\right]-s\left(\boldsymbol{x}_{r}^{k},t_{r}^{k}\right)\right)^{2},\label{eq:loss_function_r}
\end{equation}

\begin{equation}
\mathcal{L}_{i}\left(\boldsymbol{\theta}\right)=\frac{1}{N_{i}}\sum_{k=1}^{N_{i}}\left(\mathcal{I}\left[\hat{u}\left(\boldsymbol{x}_{i}^{k},t_{i}^{k};\boldsymbol{\theta}\right)\right]-f\left(\boldsymbol{x}_{i}^{k}\right)\right)^{2},\label{eq:loss_function_i}
\end{equation}

\begin{equation}
\mathcal{L}_{b}\left(\boldsymbol{\theta}\right)=\frac{1}{N_{b}}\sum_{k=1}^{N_{b}}\left(\mathcal{\mathcal{B}}\left[\hat{u}\left(\boldsymbol{x}_{b}^{k},t_{b}^{k};\boldsymbol{\theta}\right)\right]-q\left(\boldsymbol{x}_{b}^{k},t_{b}^{k}\right)\right)^{2},\label{eq:loss_function_b}
\end{equation}
and

\begin{equation}
\mathcal{L}_{d}\left(\boldsymbol{\theta}\right)=\frac{1}{N_{d}}\sum_{k=1}^{N_{d}}\left(\hat{u}\left(\boldsymbol{x}_{d}^{k},t_{d}^{k};\boldsymbol{\theta}\right)-u\left(\boldsymbol{x}_{d}^{k},t_{d}^{k}\right)\right)^{2},\label{eq:loss_function_data}
\end{equation}
where $N_{r}$, $N_{i}$, $N_{b}$ and $N_{d}$ are the number of
data points in training datasets $\left\{ \boldsymbol{x}_{r}^{k},t_{r}^{k},s\left(\boldsymbol{x}_{r}^{k},t_{r}^{k}\right)\right\} _{k=1}^{N_{r}}$,
$\left\{ \boldsymbol{x}_{i}^{k},t_{i}^{k},f\left(\boldsymbol{x}_{i}^{k}\right)\right\} _{k=1}^{N_{i}}$,
$\left\{ \boldsymbol{x}_{b}^{k},t_{b}^{k},q\left(\boldsymbol{x}_{b}^{k},t_{b}^{k}\right)\right\} _{k=1}^{N_{b}}$,
and $\left\{ \boldsymbol{x}_{d}^{k},t_{d}^{k},u\left(\boldsymbol{x}_{d}^{k},t_{d}^{k}\right)\right\} _{k=1}^{N_{d}}$,
respectively. In this context, $\boldsymbol{x}_{j}^{k}$ and $t_{j}^{k}$ for $j=r,i,b,\;\text{and}\;d$, denote discrete
spatial coordinates and discrete time points for the $k$-th data point in
each dataset, respectively. The minimizer of $\mathcal{L}\left(\boldsymbol{\Theta}\right)$
can be expressed by

\begin{equation}
\boldsymbol{\Theta}^{\star}=\mathop{\arg\min}\limits _{\boldsymbol{\Theta}}\mathcal{L}\left(\boldsymbol{\Theta}\right),\label{eq:minimize}
\end{equation}
where $\boldsymbol{\Theta}^{\star}=\left\{ \boldsymbol{\theta}^{\star},\boldsymbol{\gamma}^{\star}\right\} $
with $\boldsymbol{\theta}^{\star}$ and $\boldsymbol{\gamma}^{\star}$ 
being the optimal parameters and the estimated unknown parameters, respectively, and
$\arg\min\left(\cdot\right)$ is the arguments of the minimum. A schematic
illustration of PINNs for solving inverse problems is provided in
Fig. \ref{fig:PINN_diagram}. It is noted that the
neural network inputs, $\mathbf{x}_{1}$, ..., $\mathbf{x}_{n}$,
and $\mathbf{t}$ are collections of the spatial coordinate components
and time values, respectively, from various training datasets. Specifically,
$\mathbf{x}_{j}$ collects the $j$-th spatial
coordinate from training datasets as $\mathbf{x}_{j}=\left\{ \left(x_{j}\right)_{r}^{k}\right\} _{k=1}^{N_{r}}$
$\cup$ $\left\{ \left(x_{j}\right)_{i}^{k}\right\} _{k=1}^{N_{i}}$
$\cup$ $\left\{ \left(x_{j}\right)_{b}^{k}\right\} _{k=1}^{N_{b}}$
$\cup$ $\left\{ \left(x_{j}\right)_{d}^{k}\right\} _{k=1}^{N_{d}}$.
Similarly, $\mathbf{t}$ collects the time values from these datasets
as $\mathbf{t}=\left\{ t_{r}^{k}\right\} _{k=1}^{N_{r}}$ $\cup$
$\left\{ t_{i}^{k}\right\} _{1}^{N_{i}}$ $\cup$ $\left\{ t_{b}^{k}\right\} _{k=1}^{N_{b}}$
$\cup$ $\left\{ t_{d}^{k}\right\} _{k=1}^{N_{d}}$.
The neural network output, $\hat{\mathbf{u}}$, is formulated as $\hat{\mathbf{u}}=\left\{ \hat{u}\left(\boldsymbol{x}_{r}^{k},t_{r}^{k};\boldsymbol{\theta}\right)\right\} _{k=1}^{N_{r}}$
$\cup$ $\left\{ \hat{u}\left(\boldsymbol{x}_{i}^{k},t_{i}^{k};\boldsymbol{\theta}\right)\right\} _{k=1}^{N_{i}}$
$\cup$ $\left\{ \hat{u}\left(\boldsymbol{x}_{b}^{k},t_{b}^{k};\boldsymbol{\theta}\right)\right\} _{k=1}^{N_{b}}$
$\cup$ $\left\{ \hat{u}\left(\boldsymbol{x}_{d}^{k},t_{d}^{k};\boldsymbol{\theta}\right)\right\} _{k=1}^{N_{d}}$.

The training process of PINNs involves a range of hyperparameters
that need to be carefully tuned, including the number of hidden layers,
neurons per layer, the initialization of neural networks, the number
and distribution of data points in training datasets, types of activation
functions, optimizer settings, and the loss weights. Among these hyperparameters,
determining the appropriate values for the loss weights is crucial,
as it significantly impacts the performance of PINNs \citep{wang2021understanding,xiang2022self}.
This challenge is addressed by an adaptive weights algorithm \citep{wang2022and},
which will be briefly discussed in the following subsection.
\begin{center}
\begin{figure}[H]
\begin{centering}
\includegraphics[scale=0.75]{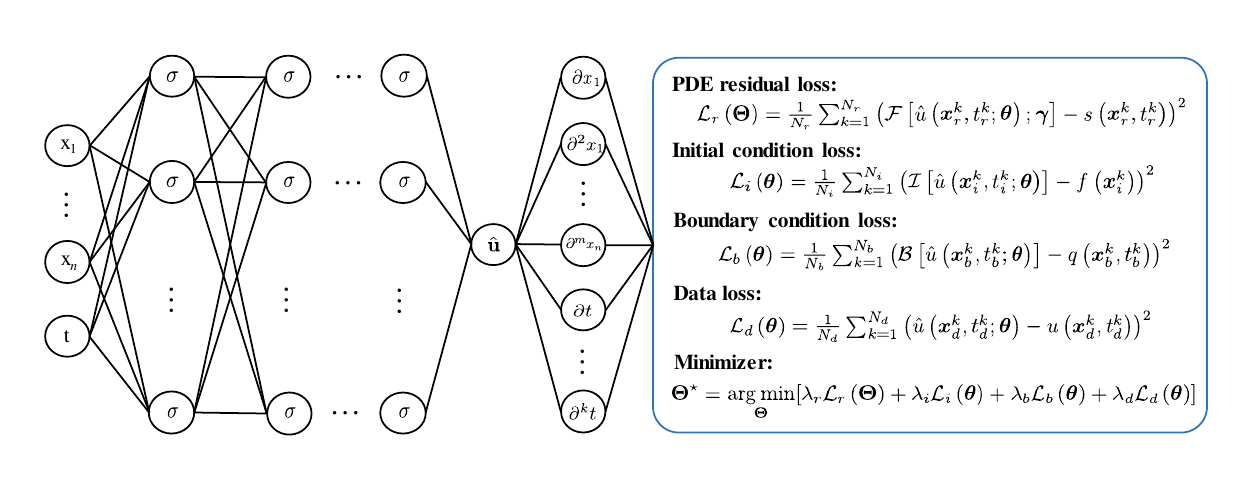}
\par\end{centering}
\caption{\label{fig:PINN_diagram}Schematic of PINNs for solving inverse problems
in PDEs.}
\end{figure}
\par\end{center}

\subsection{Adaptive weights algorithm}

Consider the minimization of the loss function in Eq. (\ref{eq:loss_function})
under the framework of a continuous-time process \citep{wang2022and},
where both $\boldsymbol{\Theta}$ and loss weights are considered
as functions of the training time $\tau$. Utilizing the gradient
descent method with an infinitesimally small learning rate $\mathrm{d}\tau$
to minimize the loss function, an update rule can be expressed by
\citep{wang2022and}

\begin{equation}
\frac{\mathrm{d}\boldsymbol{\Theta}}{\mathrm{d}\tau}=-\nabla \mathcal{L}\left(\boldsymbol{\Theta}\right),\label{eq:training_dynamics}
\end{equation}
where $\nabla$ is the gradient operator. By substituting
$\mathcal{L}\left(\boldsymbol{\Theta}\right)$ in Eq. (\ref{eq:loss_function})
into Eq. (\ref{eq:training_dynamics}), one has the NTK, denoted as
$\mathbf{K}$, that satisfies the following expression \citep{wang2022and}:

\begin{equation}
\left[\begin{array}{c}
\frac{\mathrm{d}\mathcal{F}\left[\hat{u}\left(\boldsymbol{x}_{r},t_{r};\boldsymbol{\theta}\left(\tau\right)\right);\boldsymbol{\gamma}\left(\tau\right)\right]}{\mathrm{d}\tau}\\
\frac{\mathrm{d}\mathcal{I}\left[\hat{u}\left(\boldsymbol{x}_{i},t_{i};\boldsymbol{\theta}\left(\tau\right)\right)\right]}{\mathrm{d}\tau}\\
\frac{\mathcal{\mathrm{d}B}\left[\hat{u}\left(\boldsymbol{x}_{b},t_{b};\boldsymbol{\theta}\left(\tau\right)\right)\right]}{\mathrm{d}\tau}\\
\frac{\mathrm{d}\hat{u}\left(\boldsymbol{x}_{d},t_{d};\boldsymbol{\theta}\left(\tau\right)\right)}{\mathrm{d}\tau}
\end{array}\right]=-\mathbf{K}\left[\begin{array}{c}
\mathcal{F}\left[\hat{u}\left(\boldsymbol{x}_{r},t_{r};\boldsymbol{\theta}\left(\tau\right)\right);\boldsymbol{\gamma}\left(\tau\right)\right]-s\left(\boldsymbol{x}_{r},t_{r}\right)\\
\mathcal{I}\left[\hat{u}\left(\boldsymbol{x}_{i},t_{i};\boldsymbol{\theta}\left(\tau\right)\right)\right]-f\left(\boldsymbol{x}_{i}\right)\\
\mathcal{\mathcal{B}}\left[\hat{u}\left(\boldsymbol{x}_{b},t_{b};\boldsymbol{\theta}\left(\tau\right)\right)\right]-q\left(\boldsymbol{x}_{b},t_{b}\right)\\
\hat{u}\left(\boldsymbol{x}_{d},t_{d};\boldsymbol{\theta}\left(\tau\right)\right)-u\left(\boldsymbol{x}_{d},t_{d}\right)
\end{array}\right],\label{eq:delta_theta}
\end{equation}
where

\begin{equation}
\mathbf{K}=\left[\begin{array}{c}
\frac{2\lambda_{r}}{N_{r}}\mathbf{J}_{r}\\
\frac{2\lambda_{i}}{N_{i}}\mathbf{J}_{i}\\
\frac{2\lambda_{b}}{N_{b}}\mathbf{J}_{b}\\
\frac{2\lambda_{d}}{N_{d}}\mathbf{J}_{d}
\end{array}\right]\left[\begin{array}{cccc}
\mathbf{J}_{r}^{\mathrm{T}}, & \mathbf{J}_{i}^{\mathrm{T}}, & \mathbf{J}_{b}^{\mathrm{T}}, & \mathbf{J}_{d}^{\mathrm{T}}\end{array}\right],\label{eq:NTK}
\end{equation}
with the Jacobian matrices $\mathbf{J}_{r}\in\mathbb{R}^{N_{r}\times M}$,
$\mathbf{J}_{i}\in\mathbb{R}^{N_{i}\times M}$, $\mathbf{J}_{b}\in\mathbb{R}^{N_{b}\times M}$
and $\mathbf{J}_{d}\in\mathbb{R}^{N_{d}\times M}$ of $\mathcal{F}\left[\hat{u}\left(\boldsymbol{x}_{r},t_{r};\boldsymbol{\theta}\right);\boldsymbol{\gamma}\right]\in\mathbb{R}^{N_{r}}$,
$\mathcal{I}\left[\hat{u}\left(\boldsymbol{x}_{i},t_{i};\boldsymbol{\theta}\right)\right]\in\mathbb{R}^{N_{i}}$,
$\mathcal{\mathcal{B}}\left[\hat{u}\left(\boldsymbol{x}_{b},t_{b};\boldsymbol{\theta}\right)\right]\in\mathbb{R}^{N_{b}}$
and $\hat{u}\left(\boldsymbol{x}_{d},t_{d};\boldsymbol{\theta}\right)\in\mathbb{R}^{N_{d}}$
with respect to $\boldsymbol{\Theta}\in\mathbb{R}^{M}$, respectively.
The entries of the diagonal components of $\mathbf{K}$ in the $n$-th
row and $m$-th column are expressed by

\begin{equation}
\frac{2\lambda_{r}}{N_{r}}\left[\mathbf{J}_{r}\mathbf{J}_{r}^{\mathrm{T}}\right]_{nm}=\frac{2\lambda_{r}}{N_{r}}\left\langle \frac{\mathrm{d}\mathcal{F}\left[\hat{u}\left(\boldsymbol{x}_{r}^{n},t_{r}^{n};\boldsymbol{\theta}\right);\boldsymbol{\gamma}\right]}{\mathrm{d}\boldsymbol{\Theta}},\frac{\mathrm{d}\mathcal{F}\left[\hat{u}\left(\boldsymbol{x}_{r}^{m},t_{r}^{m};\boldsymbol{\theta}\right);\boldsymbol{\gamma}\right]}{\mathrm{d}\boldsymbol{\Theta}}\right\rangle ,
\end{equation}

\begin{equation}
\frac{2\lambda_{i}}{N_{i}}\left[\mathbf{J}_{i}\mathbf{J}_{i}^{\mathrm{T}}\right]_{nm}=\frac{2\lambda_{i}}{N_{i}}\left\langle \frac{\mathrm{d}\mathcal{I}\left[\hat{u}\left(\boldsymbol{x}_{i}^{n},t_{i}^{n};\boldsymbol{\theta}\right)\right]}{\mathrm{d}\boldsymbol{\Theta}},\frac{\mathrm{d}\mathcal{I}\left[\hat{u}\left(\boldsymbol{x}_{i}^{m},t_{i}^{m};\boldsymbol{\theta}\right)\right]}{\mathrm{d}\boldsymbol{\Theta}}\right\rangle ,
\end{equation}

\begin{equation}
\frac{2\lambda_{b}}{N_{b}}\left[\mathbf{J}_{b}\mathbf{J}_{b}^{\mathrm{T}}\right]_{nm}=\frac{2\lambda_{b}}{N_{b}}\left\langle \frac{\mathcal{\mathrm{d}B}\left[\hat{u}\left(\boldsymbol{x}_{b}^{n},t_{b}^{n};\boldsymbol{\theta}\right)\right]}{\mathrm{d}\boldsymbol{\Theta}},\frac{\mathcal{\mathrm{d}B}\left[\hat{u}\left(\boldsymbol{x}_{b}^{m},t_{b}^{m};\boldsymbol{\theta}\right)\right]}{\mathrm{d}\boldsymbol{\Theta}}\right\rangle ,
\end{equation}
and

\begin{equation}
\frac{2\lambda_{d}}{N_{d}}\left[\mathbf{J}_{d}\mathbf{J}_{d}^{\mathrm{T}}\right]_{nm}=\frac{2\lambda_{d}}{N_{d}}\left\langle \frac{\mathrm{d}\hat{u}\left(\boldsymbol{x}_{d}^{n},t_{d}^{n};\boldsymbol{\theta}\right)}{\mathrm{d}\boldsymbol{\Theta}},\frac{\mathrm{d}\hat{u}\left(\boldsymbol{x}_{d}^{m},t_{d}^{m};\boldsymbol{\theta}\right)}{\mathrm{d}\boldsymbol{\Theta}}\right\rangle ,
\end{equation}
where $\left\langle \cdot,\cdot\right\rangle $ represents the inner
product. The convergence rates of $\mathcal{F}\left[\hat{u}\left(\boldsymbol{x}_{r},t_{r};\boldsymbol{\theta}\right);\boldsymbol{\gamma}\right]$,
$\mathcal{I}\left[\hat{u}\left(\boldsymbol{x}_{i},t_{i};\boldsymbol{\theta}\right)\right]$,
$\mathcal{\mathcal{B}}\left[\hat{u}\left(\boldsymbol{x}_{b},t_{b};\boldsymbol{\theta}\right)\right]$
and $\hat{u}\left(\boldsymbol{x}_{d},t_{d};\boldsymbol{\theta}\right)$
are proportional to $\frac{\lambda_{j}}{N_{j}}\mathrm{tr}\left(\mathbf{J}_{j}\mathbf{J}_{j}^{\mathrm{T}}\right)$
for $j=r,i,b,\;\text{and}\;d$, respectively, where $\mathrm{tr}\left(\cdot\right)$
denotes the trace of a matrix \citep{wang2022and}.

Based on these observations, an adaptive weights algorithm was developed
in Ref. \citep{wang2022and}. The algorithm dynamically adjusts the
loss weights by ensuring that $\frac{\lambda_{j}}{N_{j}}\mathrm{tr}\left(\mathbf{J}_{j}\mathbf{J}_{j}^{\mathrm{T}}\right)$
is equal for $j=r$, $i$, $b$, and $d$. Hence, the loss weights
can be expressed by

\begin{equation}
\lambda_{j}=\frac{N_{j}R}{\mathrm{tr}\left(\mathbf{J}_{j}\mathbf{J}_{j}^{\mathrm{T}}\right)},\quad j=r,i,b,\;\text{and}\;d,\label{eq:adaptive_weights}
\end{equation}
where
\begin{equation}
R=\frac{\mathrm{tr}\left(\mathbf{J}_{r}\mathbf{J}_{r}^{\mathrm{T}}\right)}{N_{r}}+\frac{\mathrm{tr}\left(\mathbf{J}_{i}\mathbf{J}_{i}^{\mathrm{T}}\right)}{N_{i}}+\frac{\mathrm{tr}\left(\mathbf{J}_{b}\mathbf{J}_{b}^{\mathrm{T}}\right)}{N_{b}}+\frac{\mathrm{tr}\left(\mathbf{J}_{d}\mathbf{J}_{d}^{\mathrm{T}}\right)}{N_{d}}.
\end{equation}

To facilitate the minimization process in PINNs, a two-step training
approach is often utilized \citep{he2020physics,Cuomo2022a,chen2021physics}.
It begins
with the Adam optimizer \citep{kingma2014adam}, which is an advanced variation of stochastic
gradient descent. Subsequently, the L-BFGS optimizer \citep{liu1989limited},
a quasi-Newton method, is employed. When the adaptive weights algorithm
is applied, the loss weights are updated every $1,000$ iterations
for the Adam optimizer \citep{wang2022and,ding2023physics}. Besides,
the loss weights are then fixed for the L-BFGS optimizer \citep{ding2023physics}.

\subsection{DG-PINNs for solving inverse problems}

As above-mentioned, the objective of solving an inverse problem is
to accurately estimate $\boldsymbol{\gamma}$ from the relevant PDE
and observed data. This objective is achieved by minimizing the composite
loss function in Eq. (\ref{eq:loss_function}).
It is observed that the value of $\mathcal{L}_{r}\left(\boldsymbol{\Theta}\right)$
depends on both of $\boldsymbol{\gamma}$ and $\boldsymbol{\theta}$,
while the values of $\mathcal{L}_{i}\left(\boldsymbol{\theta}\right)$,
$\mathcal{L}_{b}\left(\boldsymbol{\theta}\right)$ and $\mathcal{L}_{d}\left(\boldsymbol{\theta}\right)$
    depend only on $\boldsymbol{\theta}$. Specifically, $\mathcal{L}_{r}\left(\boldsymbol{\Theta}\right)$
ensures that $\hat{u}\left(\boldsymbol{x},t;\boldsymbol{\theta}\right)$
satisfies the PDE, $\mathcal{L}_{i}\left(\boldsymbol{\theta}\right)$
and $\mathcal{L}_{b}\left(\boldsymbol{\theta}\right)$ ensure that
$\hat{u}\left(\boldsymbol{x},t;\boldsymbol{\theta}\right)$ satisfies
the initial and boundary conditions, respectively, and $\mathcal{L}_{d}\left(\boldsymbol{\theta}\right)$
ensures that $\hat{u}\left(\boldsymbol{x},t;\boldsymbol{\theta}\right)$
fits the observed data. However, simultaneously minimizing all of these
loss terms can result in a phenomenon: $\mathcal{L}_{r}\left(\boldsymbol{\Theta}\right)$
is rapidly minimized even with an inaccurate $\boldsymbol{\gamma}$,
while $\mathcal{L}_{d}\left(\boldsymbol{\theta}\right)$ remains high
in the early training stages \citep{mao2020physics,rasht2022physics,zhou2023damageIMAC}.
Moreover, the trend of $\mathcal{L}_{i}\left(\boldsymbol{\theta}\right)$
and $\mathcal{L}_{b}\left(\boldsymbol{\theta}\right)$ is uncertain.
This phenomenon results in additional training iterations, aiming
for a more accurate $\boldsymbol{\gamma}$ and lower $\mathcal{L}\left(\boldsymbol{\Theta}\right)$.
These additional iterations are computationally costly due to the
computations of the derivatives in $\mathcal{L}_{r}\left(\boldsymbol{\Theta}\right)$
and, if existing, $\mathcal{L}_{i}\left(\boldsymbol{\theta}\right)$
and $\mathcal{L}_{b}\left(\boldsymbol{\theta}\right)$ for each iteration.

Unlike PINNs, data-driven neural networks have the loss function only with
$\mathcal{L}_{d}\left(\boldsymbol{\theta}\right)$, which can be efficiently
minimized due to the absence of derivatives in $\mathcal{L}_{d}\left(\boldsymbol{\theta}\right)$. It can be assumed that when obtaining a minimizer
$\boldsymbol{\theta}_{d}$ for $\mathcal{L}_{d}\left(\boldsymbol{\theta}\right)$
from a data-driven neural network, $\boldsymbol{\theta}_{d}$ can
approximate the optimal parameters $\boldsymbol{\theta}^{\star}$
for PINNs. When initialized with the minimizer $\boldsymbol{\theta}_{d}$,
PINNs can start with a favorable initial $\boldsymbol{\theta}$, which
can lead to quicker convergence and more accurate estimation of $\boldsymbol{\gamma}$
in the optimization process with the composite loss function in Eq.
(\ref{eq:loss_function}). Based on this logic, a novel framework,
referred to as data-guided PINNs (DG-PINNs), is proposed. 

The DG-PINNs
consist of two phases: a pre-training phase and a fine-tuning
phase. The pre-training phase is equivalent to a data-driven neural
network training methodology, where the loss function 
only contains $\mathcal{L}_{d}\left(\boldsymbol{\theta}\right)$.
The fine-tuning phase follows the PINN methodology as described in
Sec. 2.2, in which $\mathcal{L}\left(\boldsymbol{\Theta}\right)$
in Eq. (\ref{eq:loss_function}) is to be minimized. In the fine-tuning
phase, the initial parameters of the neural network, denoted by $\boldsymbol{\theta}_{0}$,
are the minimizer $\boldsymbol{\theta}_{d}$ obtained in the pre-training
phase. When $\boldsymbol{\theta}_{d} \approx \boldsymbol{\theta}^{\star}$, the fine-tuning phase not only facilitates accurate
estimation of $\boldsymbol{\gamma}$ but also requires fewer iterations
in the minimization of $\mathcal{L}\left(\boldsymbol{\Theta}\right)$.
Since the pre-training phase requires significantly lower computational
costs compared to the fine-tuning phase, DG-PINNs would offer a more
effective and computationally efficient approach for solving inverse
problems in comparison to existing PINNs. A schematic illustration
of the two-phase DG-PINNs for solving inverse problems is shown in
Fig. \ref{fig:DGPINN_diagram}.
\begin{center}
\begin{figure}[H]
\begin{centering}
\includegraphics[scale=0.75]{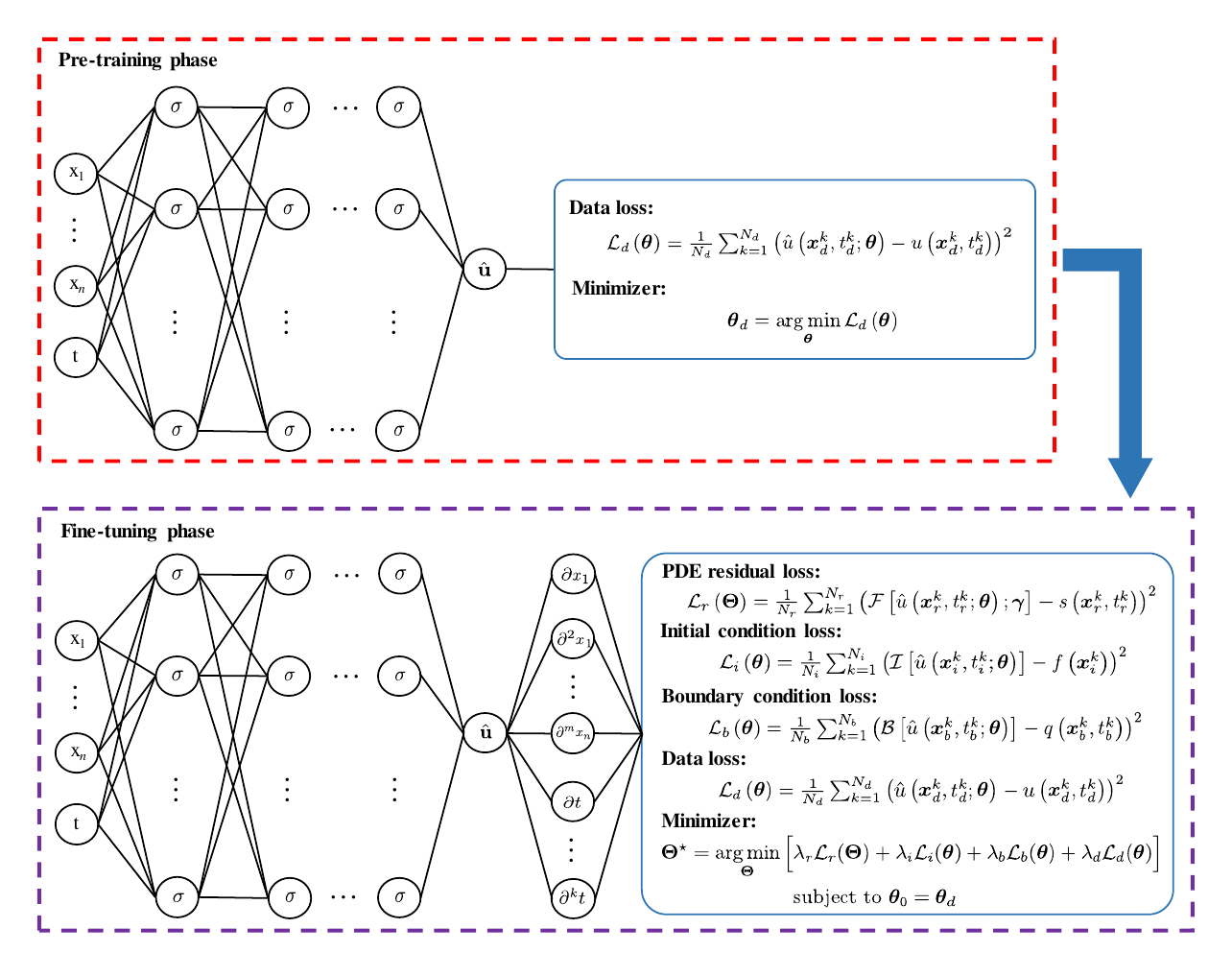}
\par\end{centering}
\caption{\label{fig:DGPINN_diagram}Schematic of DG-PINNs for solving inverse
problems in PDEs with two phases: the pre-training and fine-tuning
phases.}
\end{figure}
\par\end{center}

To efficiently execute the DG-PINNs, the Adam optimizer and L-BFGS
optimizer are used in the pre-training and fine-tuning phases, respectively.
When the pre-training phase is finished, the values of $\lambda_{r}$,
$\lambda_{i}$, $\lambda_{b}$ and $\lambda_{d}$ will be evaluated
using the adaptive weights algorithm, and they remain  constant throughout
the fine-tuning phase that follows. To evaluate the accuracy of the predictions
of DG-PINNs at the end of the fine-tuning phase, a testing dataset
$\left\{ \boldsymbol{x}_{t}^{k},t_{t}^{k},u\left(\boldsymbol{x}_{t}^{k},t_{t}^{k}\right)\right\} _{k=1}^{N_{t}}$
with the number of data points $N_{t}$ is created. The relative
$L^{2}$ error between $\mathbf{u}_{t}=\left\{ u\left(\boldsymbol{x}_{t}^{k},t_{t}^{k}\right)\right\} _{k=1}^{N_{t}}$
and the corresponding prediction $\hat{\mathbf{u}}_{t}=\left\{ \hat{u}\left(\boldsymbol{x}_{t}^{k},t_{t}^{k};\boldsymbol{\theta}\right)\right\} _{k=1}^{N_{t}}$
from DG-PINNs is utilized as a metric, which is defined as

\begin{equation}
\mathcal{R}_{t}=\frac{\left\Vert \mathbf{u}_{t}-\hat{\mathbf{u}}_{t}\right\Vert _{2}}{\left\Vert \mathbf{u}_{t}\right\Vert _{2}},\label{eq:rel2err_t}
\end{equation}
where $\left\Vert \cdot\right\Vert _{2}$ denotes the $L^{2}$ norm.

The pseudo-code for the DG-PINNs is described in Algorithm \ref{alg:DGPINNs}.
\begin{center}
\begin{algorithm}[H]
\caption{\label{alg:DGPINNs}DG-PINNs.}

\begin{algorithmic}[1]

\Require Training datasets and hyperparameters

\Ensure Optimal parameters $\boldsymbol{\theta}^{\star}$ and estimated
unknown parameters $\boldsymbol{\gamma}^{\star}$: $\boldsymbol{\Theta}^{\star}=\left\{ \boldsymbol{\theta}^{\star},\boldsymbol{\gamma}^{\star}\right\} $

\Phase{Pre-training phase}

\State Initialize neural network parameters $\boldsymbol{\theta}$

\State Set the maximum number of iterations $M_{1}$

\For{$iteration\text{ from }1\text{ to }M_{1}$}

\State Calculate the data loss $\mathcal{\mathcal{L}}_{d}\left(\boldsymbol{\theta}\right)$
using Eq. (\ref{eq:loss_function_data})

\State Update $\boldsymbol{\theta}$ using the Adam optimizer

\EndFor

\Phase{Fine-tuning phase}

\State Initialize unknown parameters $\boldsymbol{\gamma}$

\State Estimate loss weights $\lambda_{r}$, $\lambda_{i}$, $\lambda_{b}$
and $\lambda_{d}$

\State Set the maximum number of iterations $M_{2}$

\For{$iteration\text{ from }1\text{ to }M_{2}$}

\State Calculate the composite loss function $\mathcal{L}\left(\boldsymbol{\Theta}\right)$
using Eq. (\ref{eq:loss_function})

\State Update $\boldsymbol{\Theta}$ using the L-BFGS optimizer

\EndFor

\State Set $\boldsymbol{\Theta}^{\star}=\boldsymbol{\Theta}$

\State \Return $\boldsymbol{\Theta}^{\star}$

\end{algorithmic}
\end{algorithm}
\par\end{center}

\section{Numerical Investigations\label{sec:3}}

In this section, a series of numerical investigations are conducted
to validate the effectiveness, noise-robustness, and efficiency of
the DG-PINNs in solving inverse problems related to various PDEs,
including the heat equation, wave equation, Euler--Bernoulli beam
equation and Navier--Stokes equation. Throughout these numerical
investigations, three-layer fully-connected networks are used, where
each hidden layer has $100$ neurons. The activation function is the
hyperbolic tangent activation function. In this work, the initialization
of neural networks is achieved using a uniform distribution
in Ref. \citep{paszke2019pytorch}. The initialization of 
unknown parameters within PDEs is achieved using a uniform distribution
in the interval $\left[0,1\right)$.

In the numerical investigations for each inverse problem, sensitivity
analyses of the DG-PINNs
results are conducted to examine the effects of the maximum number
of iterations $M_{1}$ in the pre-training phase and the number of
data points in training datasets for the data loss. The noise-robustness of the DG-PINNs is evaluated by introducing
varying levels of white Gaussian noise to observed data. Besides,
a comparative analysis between DG-PINNs and PINNs is performed for
each inverse problem.

\subsection{One-dimensional heat equation}

The one-dimensional heat equation is a PDE that models
the temporal distribution of heat within a given region. It is a fundamental
equation in the field of heat transfer and is expressed in the dimensionless spatial-temporal
domain $\left(x,t\right)\in\left[0,1\right]\times\left[0,1\right]$
by

\begin{equation}
\frac{\partial u\left(x,t\right)}{\partial t}-\beta^{2}\frac{\partial^{2}u\left(x,t\right)}{\partial x^{2}}=0,\quad\left(x,t\right)\in\left[0,1\right]\times\left[0,1\right],\label{eq:1d_heat}
\end{equation}
where $\beta^{2}$ denotes the thermal diffusivity of the material.
This equation is subject to the following initial condition:

\begin{equation}
u\left(x,0\right)=\sin\left(10\pi x\right),\quad x\in\left[0,1\right],\label{eq:IC_heat}
\end{equation}
and boundary conditions:

\begin{equation}
u\left(0,t\right)=u\left(1,t\right)=0,\quad t\in\left[0,1\right].\label{eq:BC_heat}
\end{equation}
Applying the method of separation of variables \citep{Evans2010},
the solution $u\left(x,t\right)$ is expressed by

\begin{equation}
u\left(x,t\right)=\exp\left(-\left(10\pi\beta\right)^{2}t\right)\sin\left(10\pi x\right).\label{eq:solution_heat}
\end{equation}

Observed data for $\beta=\frac{1}{20}$ are acquired on a $201\times201$ grid evenly distributed within the spatial-temporal domain. The grid uses increments of 0.005 for both spatial and temporal dimensions. From the observed data, training
datasets $\left\{ x_{d}^{k},t_{d}^{k},u\left(x_{d}^{k},t_{d}^{k}\right)\right\} _{k=1}^{N_{d}}$
with $N_{d}=10,000$ and $\left\{ x_{r}^{k},t_{r}^{k},0\right\} _{k=1}^{N_{r}}$
with $N_{r}=2,000$ are randomly sampled in the grid, i.e., $\left(x,t\right)\in\left[0,1\right]\times\left[0,1\right]$,
$\left\{ x_{i}^{k},0,u\left(x_{i}^{k},0\right)\right\} _{k=1}^{N_{i}}$
with $N_{i}=100$ and $\left\{ x_{b}^{k},t_{b}^{k},0\right\} _{k=1}^{N_{b}}$
with $N_{b}=200$ are randomly sampled in the initial snapshot, i.e,
$t=0$ with $x\in\left[0,1\right]$, and boundaries i.e., $x=0$ or
$1$ with $t\in\left[0,1\right]$, respectively. Finally, the remaining,
unsampled observed data are grouped as a testing dataset $\left\{ x_{t}^{k},t_{t}^{k},u\left(x_{t}^{k},t_{t}^{k}\right)\right\} _{k=1}^{N_{t}}$.
These training and testing datasets are used for all investigations
in the heat equation by default. For the heat equation in Eq. (\ref{eq:1d_heat}), the PDE residual
loss is expressed by

\begin{equation}
\mathcal{L}_{r}\left(\boldsymbol{\Theta}\right)=\frac{1}{N_{r}}\sum_{k=1}^{N_{r}}\left(\frac{\partial\hat{u}\left(x_{r}^{k},t_{r}^{k};\boldsymbol{\theta}\right)}{\partial t}-\tilde{\beta}^{2}\frac{\partial^{2}\hat{u}\left(x_{r}^{k},t_{r}^{k};\boldsymbol{\theta}\right)}{\partial x^{2}}\right)^{2},\label{eq:loss_r_heat}
\end{equation}
where $\tilde{\beta}$ is the unknown parameter to be estimated.

To conduct a sensitivity analysis on the maximum number of iterations
$M_{1}$ in the pre-training phase, fifteen DG-PINN models with $M_{1}=\left\{ 2,000\right.$, $3,000$, $4,000$, $5,000$,
$6,000$, $7,000$, $8,000$, $9,000$, $10,000$, $15,000$, $20,000$,
$25,000$, $30,000$, $40,000$, $\left.50,000\right\} $, are trained.
In this sensitivity analysis, the Adam optimizer has a learning rate
of $0.001$, and the L-BFGS optimizer has a learning rate of $0.1$
with the maximum number of iterations $M_{2}$ = $10,000$.
The optimizer settings are maintained consistently for the sensitivity analysis on $M_{1}$ across the other PDEs. The accuracy of the predictions
from the fifteen DG-PINN models is evaluated by $\mathcal{R}_{t}$
as shown in Fig. \ref{fig:heat_eq_stop}(a). It can be seen that the
values of $\mathcal{R}_{t}$ remain below $6\times10^{-3}$ for all
$M_{1}$. Besides, the absolute percentage error (APE) between $\tilde{\beta}^{2}$
and $\beta^{2}$ for each DG-PINN model is shown in Fig. \ref{fig:heat_eq_stop}(b).
The maximum APE value is $4.77\%$ with $M_{1}=2000$; however a definite
correlation between the APE value and $M_{1}$ cannot be observed.
Further, since all the obtained APE values associated with the wide
range of $M_{1}$ values are considered low, DG-PINNs can solve the
inverse problem with an accurate estimation of $\beta^{2}$, which
highlights the robustness of the selection of $M_{1}$ for DG-PINNs.
\begin{center}
\begin{figure}[H]
\centering{}\subfloat[]{\centering{}\includegraphics{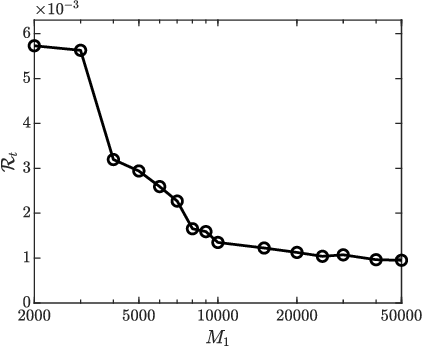}}\subfloat[]{\begin{centering}
\includegraphics{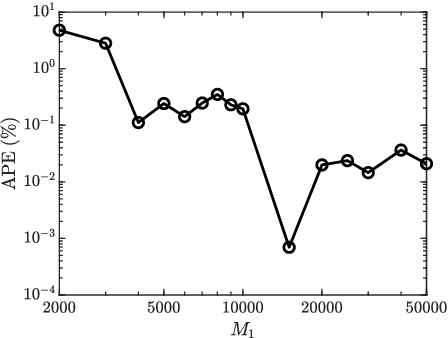}
\par\end{centering}
\centering{}}\caption{\label{fig:heat_eq_stop}Sensitivity analysis on $M_{1}$ for the
heat equation: (a) the relative $L^{2}$ errors between the testing
data and the corresponding prediction from DG-PINN models, and (b)
the absolute percentage error (APE) between $\tilde{\beta}^{2}$
and $\beta^{2}$.}
\end{figure}
\par\end{center}

To conduct a sensitivity analysis on $N_{d}$, fifteen DG-PINN models with
$\left\{ x_{d}^{k},t_{d}^{k},u\left(x_{d}^{k},t_{d}^{k}\right)\right\} _{k=1}^{N_{d}}$
with $N_{d}=\left\{500\right.$, $600$, $700$, $800$, $900$,
$1,000$, $2,000$, $3,000$, $4,000$, $5,000$, $6,000$, $7,000$,
$8,000$, $9,000$, $\left.10,000\right\}$ are trained. The Adam optimizer has a learning rate of $0.001$ with
maximum number of iterations $M_{1}$ = $20,000$, and
the L-BFGS optimizer has a learning rate of $0.1$ with the maximum number
of iterations $M_{2}$ = $10,000$. The optimizer settings are
maintained consistently for the sensitivity analysis on $N_{d}$ across
the other PDEs. The accuracy of the predictions from the fifteen DG-PINN
models is evaluated by $\mathcal{R}_{t}$ shown in Fig. \ref{fig:heat_eq_data}(a).
It can be seen that the values of $\mathcal{R}_{t}$ remain below
$4\times10^{-3}$ for all $N_{d}$ values. Besides, the APE between
$\tilde{\beta}^{2}$ and $\beta^{2}$ for each DG-PINN model is shown
in Fig. \ref{fig:heat_eq_data}(b). While the maximum APE value is
$0.18\%$ with $N_{d}=5,000$, no clear correlation between the
APE value and $N_{d}$ cannot be observed. Further, since all the
obtained APE values associated with the wide range of $N_{d}$ values
are considered low, DG-PINNs can solve the inverse problem with an
accurate estimation of $\beta^{2}$, which highlights the robustness
of the selection of $N_{d}$ for DG-PINNs.
\begin{center}
\begin{figure}[H]
\centering{}\subfloat[]{\centering{}\includegraphics{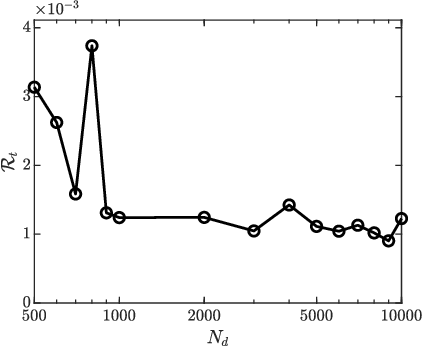}}\subfloat[]{\begin{centering}
\includegraphics{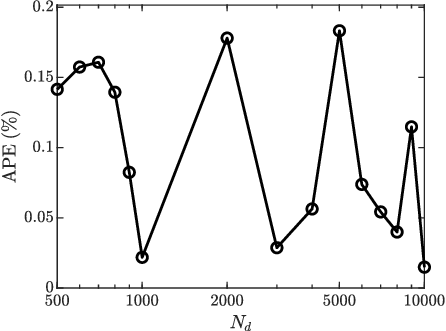}
\par\end{centering}
\centering{}}\caption{\label{fig:heat_eq_data}Sensitivity analysis on $N_{d}$ for the
heat equation: (a) the relative $L^{2}$ errors between the testing
data and the corresponding prediction from DG-PINN models, and (b)
the absolute percentage error (APE) between $\tilde{\beta}^{2}$
and $\beta^{2}$.}
\end{figure}
\par\end{center}

To investigate the noise-robustness of DG-PINNs, four DG-PINN models are trained using observed data contaminated
with white Gaussian noise at signal-to-noise ratios (SNRs)
of $40$, $35$, $30$, and $25$ dB, respectively. Training datasets
are sampled from these noise-contaminated observed data with the same
scheme as that implemented in the sensitivity analysis on $M_{1}$
for the heat equation. The Adam optimizer has a learning rate of $0.001$
with $M_{1}=50,000$, whereas the L-BFGS optimizer has a learning rate of $0.1$ with $M_{2}=20,000$. The optimizer settings are maintained consistently for the noise-robustness analysis across the other PDEs. The accuracy
of the predictions from the four DG-PINN models is evaluated by $\mathcal{R}_{t}$,
which is calculated as $1.47\times10^{-3}$, $2.25\times10^{-3}$,
$4.47\times10^{-3}$, and $9.67\times10^{-3}$ in the case of $\text{SNR}=40$,
$35$, $30$, and $25$ dB, respectively. Besides, the APE between
$\tilde{\beta}^{2}$ and $\beta^{2}$ for each DG-PINN model is calculated
as $0.16\%$, $0.14\%$, $0.24\%$, and $0.63\%
$ in the case of $\text{SNR}=40$, $35$, $30$, and $25$ dB, respectively.
These results indicate that DG-PINNs are capable of accurately solving
the inverse problem of the heat equation in the presence of noise
contamination.

To compare the efficiency of DG-PINNs and PINNs, ten independent trials
are conducted for both DG-PINNs and PINNs to solve the inverse problem
of the heat equation. In terms of optimizer settings, the Adam optimizer
has a learning rate of $0.001$ with $M_{1}=20,000$, whereas the
L-BFGS optimizer has a learning rate of $0.1$ with $M_{2}=10,000$.
The adaptive weights algorithm, outlined in Section 2.3, is utilized
to adjust loss weights every $1,000$ iterations in PINNs for the Adam
optimizer. The optimizer settings are maintained consistently for the comparative
analysis between DG-PINNs and PINNs across the other PDEs. In one of
the ten PINN trials, $\mathcal{L}\left(\boldsymbol{\Theta}\right)$
does not converge. The trial is considered an outlier and excluded
in the discussion that follows. With other convergent PINNs and DG-PINNs,
the averaged $\mathcal{R}_{t}$,	APE between $\tilde{\beta}^{2}$ and $\beta^{2}$, and training time are listed in Table \ref{tab:heat_equation}.
It is shown that $\mathcal{R}_{t}$ and APE of both PINN and DG-PINNs
are small, indicating that they both provide similarly accurate predictions.
Though the DG-PINNs have higher $\mathcal{R}_{t}$	and APE values
than PINN, the training time for DG-PINNs is significantly shorter
than that for PINNs. These results demonstrate that DG-PINNs are six times
faster than PINNs in solving the inverse problem of the heat equation,
achieving comparable accuracy in both predictions and the estimation
of $\beta^{2}$.
\begin{center}
\begin{table}[H]
\caption{\label{tab:heat_equation}Results of the comparative analysis between
DG-PINNs and PINNs for the heat equation.}

\centering{}%
\begin{tabular}{cccc}
\toprule 
Method & $\mathcal{R}_{t}$ & APE & Training time\tabularnewline
\midrule
PINN & $1.05\times10^{-3}$ & $0.05\%$ & $26.83$ min\tabularnewline
DG-PINN & $1.44\times10^{-3}$ & $0.14\%$ & $4.26$ min\tabularnewline
\bottomrule
\end{tabular}
\end{table}
\par\end{center}

\subsection{One-dimensional wave equation}

The propagation of a wave through a one-dimensional medium is a fundamental
phenomenon observed in various physical systems. This process is governed
by the one-dimensional wave equation, a crucial mathematical model
in the field of wave propagation. Formally, within the dimensionless
spatial-temporal domain $\left(x,t\right)\in\left[0,1\right]\times\left[0,1\right]$,
the equation is expressed by

\begin{equation}
\frac{\partial^{2}u\left(x,t\right)}{\partial t^{2}}-c^{2}\frac{\partial^{2}u\left(x,t\right)}{\partial x^{2}}=0,\quad\left(x,t\right)\in\left[0,1\right]\times\left[0,1\right],\label{eq:1d_wave}
\end{equation}
where $c$ denotes the wave speed. This equation is subject to initial conditions:

\begin{equation}
\begin{cases}
u\left(x,0\right)=\sin\left(\pi x\right)+\frac{1}{2}\sin\left(4\pi x\right)\\
\frac{\partial u\left(x,0\right)}{\partial t}=0
\end{cases},\quad x\in\left[0,1\right],\label{eq:IC_wave}
\end{equation}
and boundary conditions:

\begin{equation}
u\left(0,t\right)=u\left(1,t\right)=0,\quad t\in\left[0,1\right].\label{eq:BC_wave}
\end{equation}
By applying the method of separation of variables \citep{Evans2010},
the solution $u\left(x,t\right)$ can be analytically expressed by

\begin{equation}
u\left(x,t\right)=\sin\left(\pi x\right)\cos\left(c\pi t\right)+\frac{1}{2}\sin\left(4\pi x\right)\cos\left(4c\pi t\right).\label{eq:solution_wave}
\end{equation}

Observed data of the wave equation with $c=2$ are acquired on a $201\times201$ grid uniformly distributed within the domain $\left(x,t\right)\in\left[0,1\right]\times\left[0,1\right]$. The acquisition process for the training and testing datasets is the same as that implemented in the sensitivity analysis on $M_{1}$
for the heat equation in Sec. 3.1. These training and testing datasets
are used for all investigations in the wave equation by default. The PDE residual loss $\mathcal{L}_{r}\left(\boldsymbol{\Theta}\right)$
for solving the inverse problem in the wave equation is expressed
by

\begin{equation}
\mathcal{L}_{r}\left(\boldsymbol{\Theta}\right)=\frac{1}{N_{r}}\sum_{k=1}^{N_{r}}\left(\frac{\partial^{2}\hat{u}\left(x_{r}^{k},t_{r}^{k};\boldsymbol{\theta}\right)}{\partial t^{2}}-\tilde{c}^{2}\frac{\partial^{2}\hat{u}\left(x_{r}^{k},t_{r}^{k};\boldsymbol{\theta}\right)}{\partial x^{2}}\right)^{2},\label{eq:loss_r_wave}
\end{equation}
where $\tilde{c}$ is the unknown parameter to be estimated. Two initial
condition losses $\mathcal{L}_{i1}\left(\boldsymbol{\theta}\right)$
and $\mathcal{L}_{i2}\left(\boldsymbol{\theta}\right)$ are defined,
which are expressed by

\begin{equation}
\mathcal{L}_{i1}\left(\boldsymbol{\theta}\right)=\frac{1}{N_{i}}\sum_{k=1}^{N_{i}}\left(\hat{u}\left(x_{i}^{k},0;\boldsymbol{\theta}\right)-\sin\left(\pi x_{i}^{k}\right)-\frac{1}{2}\sin\left(4\pi x_{i}^{k}\right)\right)^{2},\label{eq:loss_i_wave-1}
\end{equation}
and

\begin{equation}
\mathcal{L}_{i2}\left(\boldsymbol{\theta}\right)=\frac{1}{N_{i}}\sum_{k=1}^{N_{i}}\left(\frac{\partial\hat{u}\left(x_{i}^{k},0;\boldsymbol{\theta}\right)}{\partial t}\right)^{2},\label{eq:loss_i_wave-2}
\end{equation}
respectively. Boundary condition loss $\mathcal{L}_{b}\left(\boldsymbol{\theta}\right)$
and the data loss $\mathcal{L}_{d}\left(\boldsymbol{\theta}\right)$
can be calculated using Eqs. (\ref{eq:loss_function_b}) and (\ref{eq:loss_function_data}),
respectively. The composite loss function for the wave equation can
be expressed by

\begin{equation}
\mathcal{L}\left(\boldsymbol{\Theta}\right)=\lambda_{r}\mathcal{L}_{r}\left(\boldsymbol{\Theta}\right)+\lambda_{i1}\mathcal{L}_{i1}\left(\boldsymbol{\theta}\right)+\lambda_{i2}\mathcal{L}_{i2}\left(\boldsymbol{\theta}\right)+\lambda_{b}\mathcal{L}_{b}\left(\boldsymbol{\theta}\right)+\lambda_{d}\mathcal{L}_{d}\left(\boldsymbol{\theta}\right).\label{eq:loss_wave}
\end{equation}

To conduct a sensitivity analysis on $M_{1}$ for the wave
equation, fifteen DG-PINN models are trained with same $M_{1}$ values as those
specified in the sensitivity analysis on $M_{1}$ for the heat equation
in Sec. 3.1. The accuracy of the predictions from the fifteen DG-PINN
models is evaluated by $\mathcal{R}_{t}$ shown in Fig. \ref{fig:wave_eq_stop}(a).
It can be seen that the values of $\mathcal{R}_{t}$ remain below
$3\times10^{-3}$ for all $M_{1}$. Besides, the APE between $\tilde{c}^{2}$
and $c^{2}$ for each DG-PINN model is shown in Fig. \ref{fig:wave_eq_stop}(b).
The maximum APE value is $0.027\%$ with $M_{1}=2,000$; however, no indicative correlation between the APE value and $M_{1}$ can be
observed. Further, since all the obtained APE values associated with
the wide range of $M_{1}$ values are considered low, DG-PINNs can
solve the inverse problem with an accurate estimation of $c^{2}$,
which highlights the robustness of the selection of $M_{1}$ for DG-PINNs.
\begin{center}
\begin{figure}[H]
\centering{}\subfloat[]{\centering{}\includegraphics{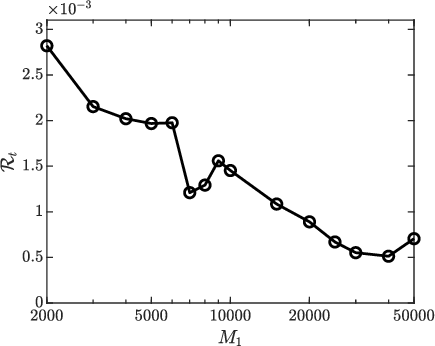}}\subfloat[]{\begin{centering}
\includegraphics{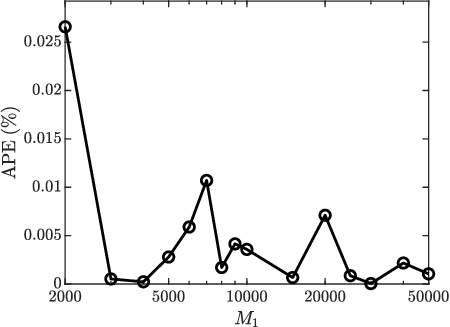}
\par\end{centering}
\centering{}}\caption{\label{fig:wave_eq_stop}Sensitivity analysis on $M_{1}$ for the
wave equation: (a) the relative $L^{2}$ errors between the testing
data and the corresponding predictions from the DG-PINN models, and
(b) the absolute percentage error (APE) between $\tilde{c}^{2}$
and $c^{2}$.}
\end{figure}
\par\end{center}

\begin{figure}[H]
\centering{}\subfloat[]{\centering{}\includegraphics{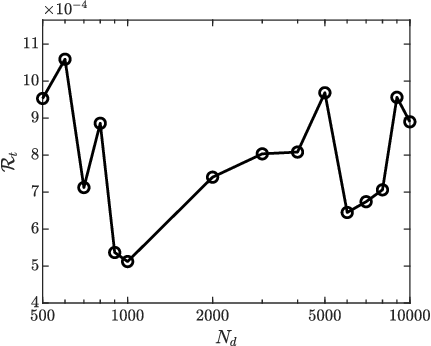}}\subfloat[]{\begin{centering}
\includegraphics{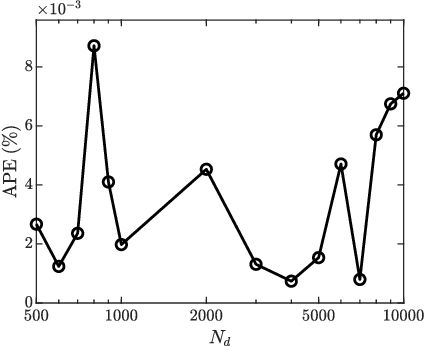}
\par\end{centering}
\centering{}}\caption{\label{fig:wave_eq_data}Sensitivity analysis on $N_{d}$ for the
wave equation: (a) the relative $L^{2}$ errors between the testing
data and the corresponding predictions of the DG-PINN models, and
(b) the absolute percentage error (APE) between $\tilde{c}^{2}$
and $c^{2}$.}
\end{figure}

To conduct a sensitivity analysis on $N_{d}$, fifteen DG-PINN models are trained using 
$\left\{ x_{d}^{k},t_{d}^{k},u\left(x_{d}^{k},t_{d}^{k}\right)\right\} _{k=1}^{N_{d}}$
for $\mathcal{L}_{d}\left(\boldsymbol{\theta}\right)$ in the wave equation, with $N_{d}$ values matching those used in the sensitivity analysis for the heat equation. The accuracy of the predictions from the fifteen
DG-PINN models is evaluated by $\mathcal{R}_{t}$ shown in Fig. \ref{fig:wave_eq_data}(a).
It can be seen that the values of $\mathcal{R}_{t}$ remain below
$1.1\times10^{-3}$. Besides, the APE between $\tilde{c}^{2}$ and
$c^{2}$ for each DG-PINN model is shown in Fig. \ref{fig:wave_eq_data}(b).
While the maximum APE value is $0.009\%$ with $N_{d}=800$, an indicative
correlation between the APE value and $N_{d}$ cannot be observed.
Further, since all the obtained APE values associated with the wide
range of $N_{d}$ values are considered low, DG-PINNs can solve the
inverse problem with an accurate estimation of $c^{2}$, which highlights
the robustness of the selection of $N_{d}$ for DG-PINNs.

To investigate the noise-robustness of DG-PINNs for the wave equation, four DG-PINN models are trained using observed data contaminated
with white Gaussian noise at SNRs of $40$, $35$, $30$,
and $25$ dB, respectively. Training datasets are sampled from these
noise-contaminated observed data. The accuracy of the predictions
from the four DG-PINN models is evaluated using $\mathcal{R}_{t}$, which
is calculated as $6.59\times10^{-4}$, $1.16\times10^{-3}$, $2.00\times10^{-3}$,
and $3.63\times10^{-3}$ for SNRs of $40$, $35$, $30$, and $25$
dB, respectively. Besides, the APE between $\tilde{c}^{2}$ and $c^{2}$
for each DG-PINN model is calculated as $0.002\%$, $0.003\%$, $0.007\%$,
and $0.006\%$ in the case of $\text{SNR}=40$, $35$, $30$, and $25$ dB, respectively.
These results indicate that DG-PINNs are capable of accurately solving
the inverse problem of the wave equation in the presence of noise
contamination.

To evaluate the efficiency of PINNs and DG-PINNs, both approaches
are applied in ten independent trials to solve the inverse problem
of the wave equation. The averaged $\mathcal{R}_{t}$, APE between
$\tilde{c}^{2}$ and $c^{2}$, and training time are listed in Table
\ref{tab:wave_equation}. It is shown that $\mathcal{R}_{t}$ for
DG-PINNs is lower than that for PINNs, indicating that DG-PINNs provide
more accurate predictions compared to PINNs. Regarding the APE between $\tilde{c}^{2}$
and $c^{2}$, both DG-PINNs and PINNs exhibit high accuracy with
significantly small APE values. Besides, the computational efficiency of
DG-PINNs is significantly higher, as evidenced by markedly reduced
training times compared to PINNs. These results demonstrate that DG-PINNs
are six times faster than PINNs in solving the inverse problem of
the wave equation, achieving comparable accuracy in both predictions and
the estimation of $c^{2}$.
\begin{center}
\begin{table}[H]
\caption{\label{tab:wave_equation}Results of the comparative analysis between
DG-PINNs and PINNs for the wave equation.}

\centering{}%
\begin{tabular}{cccc}
\toprule 
Method & $\mathcal{R}_{t}$ & APE & Training time\tabularnewline
\midrule
PINN & $1.3\times10^{-3}$ & $0.003\%$ & $28.57$ min\tabularnewline
DG-PINN & $7.49\times10^{-4}$ & $0.003\%$ & $4.57$ min\tabularnewline
\bottomrule
\end{tabular}
\end{table}
\par\end{center}

\subsection{Euler--Bernoulli beam equation}

The Euler-Bernoulli beam equation, a fundamental principle in the
field of structural analysis and engineering, describes the flexural
motion of beam \citep{hagedorn2007vibrations}. When considering a
homogeneous beam undergoing undamped free flexural vibration, the governing equation for its flexural motion within
the dimensionless spatial-temporal domain $\left(x,t\right)\in\left[0,1\right]\times\left[0,1\right]$
is expressed by
\begin{equation}
\frac{\partial^{2}u\left(x,t\right)}{\partial t^{2}}+\alpha^{2}\frac{\partial^{4}u\left(x,t\right)}{\partial x^{4}}=0,\quad\left(x,t\right)\in\left[0,1\right]\times\left[0,1\right],\label{eq:Euler_beam_free}
\end{equation}
where $\alpha$ is a parameter related to the material property of the beam.
This equation is subject to initial conditions:

\begin{equation}
\begin{cases}
u\left(x,0\right)=\sin\left(\pi x\right)\\
\frac{\partial u\left(x,0\right)}{\partial t}=0
\end{cases},\quad x\in\left[0,1\right],\label{eq:IC_beam}
\end{equation}
and boundary conditions:

\begin{equation}
\begin{cases}
u\left(0,t\right)=u\left(1,t\right)=0\\
\frac{\partial^{2}u\left(0,t\right)}{\partial x^{2}}=\frac{\partial^{2}u\left(1,t\right)}{\partial x^{2}}=0
\end{cases},\quad t\in\left[0,1\right].\label{eq:BC_beam}
\end{equation}
By applying the method of separation of variables \citep{Evans2010},
the solution $u\left(x,t\right)$ can be analytically expressed by

\begin{equation}
u\left(x,t\right)=\sin\left(\pi x\right)\cos\left(\alpha\pi^{2}t\right).\label{eq:solution_beam}
\end{equation}

Observed data for the beam equation with $\alpha=1$ are acquired
on a $201\times201$ grid uniformly
distributed within the domain $\left(x,t\right)\in\left[0,1\right]\times\left[0,1\right]$.
The acquisition of datasets for training and testing is the same as the
approach implemented in the sensitivity analysis on $M_{1}$ for the
heat equation in Sec. 3.1. These training and testing datasets are
used for all investigations in the beam equation by default. The PDE residual loss $\mathcal{L}_{r}\left(\boldsymbol{\Theta}\right)$
for the beam equation is expressed by

\begin{equation}
\mathcal{L}_{r}\left(\boldsymbol{\Theta}\right)=\frac{1}{N_{r}}\sum_{k=1}^{N_{r}}\left(\frac{\partial^{2}\hat{u}\left(x_{r}^{k},t_{r}^{k};\boldsymbol{\theta}\right)}{\partial t^{2}}+\tilde{\alpha}^{2}\frac{\partial^{4}\hat{u}\left(x_{r}^{k},t_{r}^{k};\boldsymbol{\theta}\right)}{\partial x^{4}}\right)^{2},\label{eq:loss_r_beam}
\end{equation}
where $\tilde{\alpha}$ is the unknown parameter to be estimated.
Two initial condition losses $\mathcal{L}_{i1}\left(\boldsymbol{\theta}\right)$
and $\mathcal{L}_{i2}\left(\boldsymbol{\theta}\right)$ are defined,
which are expressed by

\begin{equation}
\mathcal{L}_{i1}\left(\boldsymbol{\theta}\right)=\frac{1}{N_{i}}\sum_{k=1}^{N_{i}}\left(\hat{u}\left(x_{i}^{k},0;\boldsymbol{\theta}\right)-\sin\left(\pi x_{i}^{k}\right)\right)^{2},\label{eq:loss_i_beam-1}
\end{equation}
and

\begin{equation}
\mathcal{L}_{i2}\left(\boldsymbol{\theta}\right)=\frac{1}{N_{i}}\sum_{k=1}^{N_{i}}\left(\frac{\partial\hat{u}\left(x_{i}^{k},0;\boldsymbol{\theta}\right)}{\partial t}\right)^{2},\label{eq:loss_i_beam-2}
\end{equation}
respectively. Two boundary condition losses $\mathcal{L}_{b1}\left(\boldsymbol{\theta}\right)$
and $\mathcal{L}_{b2}\left(\boldsymbol{\theta}\right)$ are defined,
which are expressed by

\begin{equation}
\mathcal{L}_{b1}\left(\boldsymbol{\theta}\right)=\frac{1}{N_{b}}\sum_{k=1}^{N_{b}}\left(\hat{u}\left(x_{b}^{k},t_{b}^{k};\boldsymbol{\theta}\right)\right)^{2},\label{eq:loss_b_beam-1}
\end{equation}
and

\begin{equation}
\mathcal{L}_{b2}\left(\boldsymbol{\theta}\right)=\frac{1}{N_{b}}\sum_{k=1}^{N_{b}}\left(\frac{\partial^{2}\hat{u}\left(x_{b}^{k},t_{b}^{k};\boldsymbol{\theta}\right)}{\partial x^{2}}\right)^{2}.\label{eq:loss_b_beam-2}
\end{equation}
The data loss $\mathcal{L}_{d}\left(\boldsymbol{\theta}\right)$ can
be calculated using Eq. (\ref{eq:loss_function_data}). The composite
loss function for the beam equation can be expressed by
\begin{equation}
\mathcal{L}\left(\boldsymbol{\Theta}\right)=\lambda_{r}\mathcal{L}_{r}\left(\boldsymbol{\Theta}\right)+\lambda_{i1}\mathcal{L}_{i1}\left(\boldsymbol{\theta}\right)+\lambda_{i2}\mathcal{L}_{i2}\left(\boldsymbol{\theta}\right)+\lambda_{b1}\mathcal{L}_{b1}\left(\boldsymbol{\theta}\right)+\lambda_{b2}\mathcal{L}_{b2}\left(\boldsymbol{\theta}\right)+\lambda_{d}\mathcal{L}_{d}\left(\boldsymbol{\theta}\right).\label{eq:loss_beam}
\end{equation}

To conduct a sensitivity analysis on $M_{1}$ for the beam
equation, fifteen DG-PINN models are trained with same $M_{1}$ values as those
specified in the sensitivity analysis on $M_{1}$ for the heat equation
in Sec. 3.1. The accuracy of the predictions from the fifteen DG-PINN
models is evaluated by $\mathcal{R}_{t}$ shown in Fig. \ref{fig:beam_eq_stop}(a).
It can be seen that the values of $\mathcal{R}_{t}$ remain below
$3\times10^{-4}$ for all $M_{1}$. Besides, the APE between $\tilde{\alpha}^{2}$
and $\alpha^{2}$ for each DG-PINN model is shown in Fig. \ref{fig:beam_eq_stop}(b).
The maximum APE value is $0.034\%$ with $M_{1}=10,000$; however, no indicative correlation between the APE value and $M_{1}$ can be
observed. Further, since all the obtained APE values associated with
the wide range of $M_{1}$ values are considered low, DG-PINNs can
solve the inverse problem with an accurate estimation of $\alpha^{2}$,
which highlights the robustness of the selection of $M_{1}$ for DG-PINNs.
\begin{center}
\begin{figure}[H]
\centering{}\subfloat[]{\centering{}\includegraphics{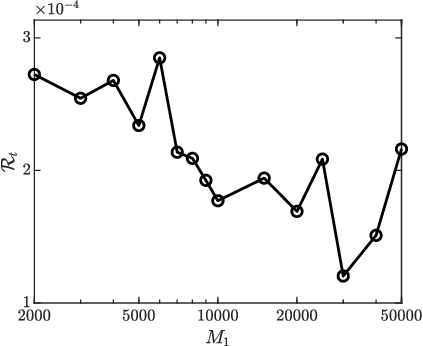}}\subfloat[]{\begin{centering}
\includegraphics{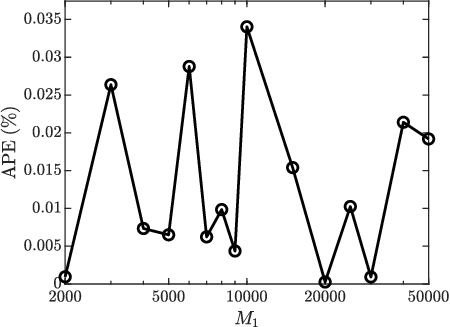}
\par\end{centering}
\centering{}}\caption{\label{fig:beam_eq_stop}Sensitivity analysis on $M_{1}$ for the
beam equation: (a) the relative $L^{2}$ errors between the testing
data and the corresponding prediction of the DG-PINN models, and (b)
the absolute percentage error (APE) between $\tilde{\alpha}^{2}$
and $\alpha^{2}$.}
\end{figure}
\par\end{center}

\begin{figure}[H]
\centering{}\subfloat[]{\centering{}\includegraphics{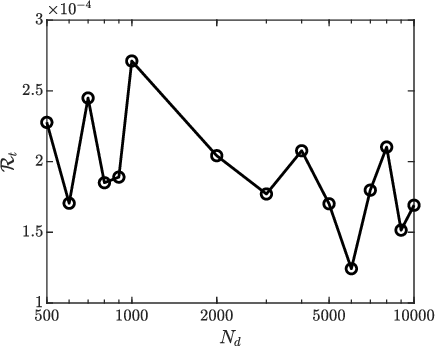}}\subfloat[]{\begin{centering}
\includegraphics{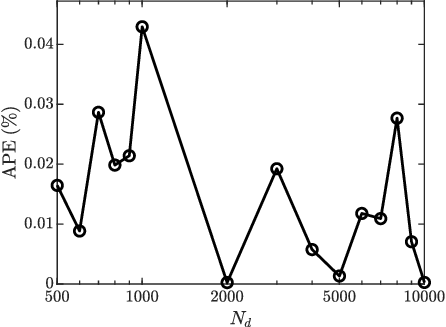}
\par\end{centering}
\centering{}}\caption{\label{fig:beam_eq_data}Sensitivity analysis of $N_{d}$ for the
beam equation: (a) the relative $L^{2}$ errors between the testing
data and the corresponding prediction of the DG-PINN models, and (b)
the absolute percentage error (APE) between $\tilde{\alpha}^{2}$
and $\alpha^{2}$.}
\end{figure}

To conduct a sensitivity analysis on $N_{d}$, fifteen DG-PINN models are trained using 
$\left\{ x_{d}^{k},t_{d}^{k},u\left(x_{d}^{k},t_{d}^{k}\right)\right\} _{k=1}^{N_{d}}$
for $\mathcal{L}_{d}\left(\boldsymbol{\theta}\right)$ in the beam equation, with $N_{d}$ values matching those used in the sensitivity analysis for the heat equation. The accuracy of the predictions
from the fifteen DG-PINN models is evaluated by $\mathcal{R}_{t}$
shown in Fig. \ref{fig:beam_eq_data}(a). It can be seen that the
values of $\mathcal{R}_{t}$ remain below $3\times10^{-4}$. Besides,
the APE between $\tilde{\alpha}^{2}$ and $\alpha^{2}$ for each DG-PINN
model is shown in Fig. \ref{fig:beam_eq_data}(b). The maximum
APE value is $0.043\%$ with $N_{d}=1,000$; however, an indicative correlation
between the APE value and $N_{d}$ cannot be observed. Further, since
all the obtained APE values associated with the wide range of $N_{d}$
values are considered low, DG-PINNs can solve the inverse problem
with an accurate estimation of $\alpha^{2}$, which highlights the
robustness of the selection of $N_{d}$ for DG-PINNs.

To investigate the noise-robustness of DG-PINNs for the beam equation, four DG-PINN models are trained using observed data contaminated
with white Gaussian noise at SNRs of $40$, $35$, $30$,
and $25$ dB, respectively. Training datasets are sampled from these
noise-contaminated observed data. The accuracy of the predictions from
the four DG-PINN models is evaluated using $\mathcal{R}_{t}$, which
is calculated as $4.89\times10^{-4}$, $8.40\times10^{-4}$, $1.50\times10^{-3}$
and $3.49\times10^{-3}$ for SNRs of $40$, $35$, $30$, and $25$
dB, respectively. Besides, the APE between $\tilde{\alpha}^{2}$ and
$\alpha^{2}$ for each DG-PINN model is calculated as $0.056\%$,
$0.065\%$, $0.109\%$, and $0.852\%
$ in the case of $\text{SNR}=40$, $35$, $30$, and $25$ dB, respectively.
These results indicate that DG-PINNs are capable of accurately solving
the inverse problem of the beam equation in the presence of noise
contamination.

To evaluate the efficiency of PINNs and DG-PINNs, both approaches
are applied in ten independent trials to solve the inverse problem
of the beam equation. The averaged $\mathcal{R}_{t}$, APE between
$\tilde{\alpha}^{2}$ versus $\alpha^{2}$, and training time are listed in Table \ref{tab:beam_equation}. It is
shown that $\mathcal{R}_{t}$ for DG-PINNs is lower than that for PINNs, indicating that DG-PINNs provide more accurate predictions
compared to PINNs. Regarding the APE between $\tilde{\alpha}^{2}$ and $\alpha^{2}$,
both DG-PINNs and PINNs exhibit high accuracy with small APE values,
but PINNs provide a more accurate estimation of $\alpha^{2}$. Besides,
the training time for DG-PINNs is significantly shorter than for PINNs.
These results demonstrate that DG-PINNs are six times faster than
PINNs in solving the inverse problem of the beam equation, achieving comparable
accuracy in both predictions and the estimation of $\alpha^{2}$.
\begin{center}
\begin{table}[H]
\caption{\label{tab:beam_equation}Results of the comparative analysis between
DG-PINNs and PINNs for the beam equation.}

\centering{}%
\begin{tabular}{cccc}
\toprule 
Method & $\mathcal{R}_{t}$ & APE & Training time\tabularnewline
\midrule
PINN & $4.10\times10^{-4}$ & $0.01\%$ & $38.04$ min\tabularnewline
DG-PINN & $2.62\times10^{-4}$ & $0.10\%$ & $6.43$ min\tabularnewline
\bottomrule
\end{tabular}
\end{table}
\par\end{center}

\subsection{Navier--Stokes equation}

The Navier--Stokes equations for an incompressible fluid describe
the motion of fluid substances, such as liquids and gases. These equations
are fundamental in the field of fluid dynamics. For an incompressible
fluid, the Navier--Stokes equations in two dimensions can be expressed
by
\begin{equation}
\begin{cases}
\frac{\partial u}{\partial t}+\beta_{1}\left(u\frac{\partial u}{\partial x}+v\frac{\partial u}{\partial y}\right)=-\frac{\partial p}{\partial x}+\beta_{2}\left(\frac{\partial^{2}u}{\partial x^{2}}+\frac{\partial^{2}u}{\partial y^{2}}\right)\\
\frac{\partial v}{\partial t}+\beta_{1}\left(u\frac{\partial v}{\partial x}+v\frac{\partial v}{\partial y}\right)=-\frac{\partial p}{\partial y}+\beta_{2}\left(\frac{\partial^{2}v}{\partial x^{2}}+\frac{\partial^{2}v}{\partial y^{2}}\right),\\
\frac{\partial u}{\partial x}+\frac{\partial v}{\partial y}=0
\end{cases}\label{eq:NS_eq}
\end{equation}
where $u\left(x,y,t\right)$ and $v\left(x,y,t\right)$ denote the
$x$- and $y$-component of the velocity field, respectively, and
$p\left(x,y,t\right)$ denotes the pressure field.

The prototype problem of incompressible flow past a circular cylinder
same as one in Ref. \citep{raissi2019physics} is used, where $\beta_{1}=1$
and $\beta_{2}=0.01$. In solving the inverse problem of the Navier--Stokes
equation, estimating parameter $\beta_{1}$ and $\beta_{2}$ as well
as the pressure $p\left(x,y,t\right)$ are interested. A high-resolution
dataset in domain $\left(x,y,t\right)\in\left[1,8\right]\times\left[-2,2\right]\times\left[0,7\right]$.
Training datasets $\left\{ x_{d}^{k},y_{d}^{k},t_{d}^{k},u\left(x_{d}^{k},y_{d}^{k},t_{d}^{k}\right),v\left(x_{d}^{k},y_{d}^{k},t_{d}^{k}\right)\right\} _{k=1}^{N_{d}}$
with $N_{d}=10,000$ and $\left\{ x_{r}^{k},y_{r}^{k},t_{r}^{k},0\right\} _{k=1}^{N_{r}}$
with $N_{r}=2,000$ are randomly sampled in the interior of the grid. Other unused observed data are served as a testing dataset $\left\{ x_{t}^{k},y_{t}^{k},t_{t}^{k},u\left(x_{t}^{k},y_{t}^{k},t_{t}^{k}\right),v\left(x_{t}^{k},y_{t}^{k},t_{t}^{k}\right)\right\} _{k=1}^{N_{t}}$.

Assume outputs of a neural network $\hat{u}\left(x,y,t;\boldsymbol{\theta}\right)$,
$\hat{v}\left(x,y,t;\boldsymbol{\theta}\right)$ and $\hat{p}\left(x,y,t;\boldsymbol{\theta}\right)$
approximate $u\left(x,y,t\right)$, $v\left(x,y,t\right)$ and $p\left(x,y,t\right)$.
The residuals $f\left(x,y,t;\boldsymbol{\Theta}\right)$, $g\left(x,y,t;\boldsymbol{\Theta}\right)$,
and $h\left(x,y,t;\boldsymbol{\theta}\right)$ can be given by

\begin{equation}
\begin{cases}
f\left(x,y,t;\boldsymbol{\Theta}\right)=\frac{\partial\hat{u}}{\partial t}+\tilde{\beta_{1}}\left(\hat{u}\frac{\partial\hat{u}}{\partial x}+\hat{v}\frac{\partial\hat{u}}{\partial y}\right)+\frac{\partial\hat{p}}{\partial x}-\tilde{\beta_{2}}\left(\frac{\partial^{2}\hat{u}}{\partial x^{2}}+\frac{\partial^{2}\hat{u}}{\partial y^{2}}\right)\\
g\left(x,y,t;\boldsymbol{\Theta}\right)=\frac{\partial\hat{v}}{\partial t}+\tilde{\beta_{1}}\left(\hat{u}\frac{\partial\hat{v}}{\partial x}+\hat{v}\frac{\partial\hat{v}}{\partial y}\right)+\frac{\partial\hat{p}}{\partial y}-\tilde{\beta_{2}}\left(\frac{\partial^{2}\hat{v}}{\partial x^{2}}+\frac{\partial^{2}\hat{v}}{\partial y^{2}}\right),\\
h\left(x,y,t;\boldsymbol{\theta}\right)=\frac{\partial\hat{u}\left(x_{r}^{k},y_{r}^{k},t_{r}^{k}\right)}{\partial x}+\frac{\partial\hat{v}\left(x_{r}^{k},y_{r}^{k},t_{r}^{k}\right)}{\partial y}
\end{cases}\label{eq:NS_residual}
\end{equation}
where $\tilde{\beta_{1}}$ and $\tilde{\beta_{2}}$ are the unknown
parameters to be estimated for the inverse problem in the Navier--Stokes
equation. The loss function for the pre-training phase of DG-PINNs
is formulated by

\begin{equation}
\mathcal{L}_{d}\left(\boldsymbol{\theta}\right)=\mathcal{L}_{d1}\left(\boldsymbol{\theta}\right)+\mathcal{L}_{d2}\left(\boldsymbol{\theta}\right),\label{eq:data_loss_NS}
\end{equation}
where

\begin{equation}
\mathcal{L}_{d1}\left(\boldsymbol{\theta}\right)=\frac{1}{N_{d}}\sum_{k=1}^{N_{d}}\left(\hat{u}\left(x_{d}^{k},y_{d}^{k},t_{d}^{k}\text{;}\boldsymbol{\theta}\right)-u\left(x_{d}^{k},y_{d}^{k},t_{d}^{k}\right)\right)^{2},
\end{equation}
and

\begin{equation}
\mathcal{L}_{d2}\left(\boldsymbol{\theta}\right)=\frac{1}{N_{d}}\sum_{k=1}^{N_{d}}\left(\hat{v}\left(x_{d}^{k},y_{d}^{k},t_{d}^{k};\boldsymbol{\theta}\right)-v\left(x_{d}^{k},y_{d}^{k},t_{d}^{k}\right)\right)^{2}.
\end{equation}
Then, the loss function for the fine-tuning phase of DG-PINNs is formulated
by

\begin{equation}
\mathcal{L}\left(\boldsymbol{\Theta}\right)=\lambda_{r1}\mathcal{L}_{r1}\left(\boldsymbol{\Theta}\right)+\lambda_{r2}\mathcal{L}_{r2}\left(\boldsymbol{\Theta}\right)+\lambda_{r3}\mathcal{L}_{r3}\left(\boldsymbol{\theta}\right)+\lambda_{d1}\mathcal{L}_{d1}\left(\boldsymbol{\theta}\right)+\lambda_{d2}\mathcal{L}_{d2}\left(\boldsymbol{\theta}\right),\label{eq:loss_NS}
\end{equation}
where

\begin{equation}
\mathcal{L}_{r1}\left(\boldsymbol{\Theta}\right)=\frac{1}{N_{r}}\sum_{k=1}^{N_{r}}\left(f\left(x_{r}^{k},y_{r}^{k},t_{r}^{k};\boldsymbol{\Theta}\right)\right)^{2},\label{eq:loss_r1_NS}
\end{equation}

\begin{equation}
\mathcal{L}_{r2}\left(\boldsymbol{\Theta}\right)=\frac{1}{N_{r}}\sum_{k=1}^{N_{r}}\left(g\left(x_{r}^{k},y_{r}^{k},t_{r}^{k};\boldsymbol{\Theta}\right)\right)^{2},\label{eq:loss_r2_NS}
\end{equation}
and

\begin{equation}
\mathcal{L}_{r3}\left(\boldsymbol{\theta}\right)=\frac{1}{N_{r}}\sum_{k=1}^{N_{r}}\left(h\left(x_{r}^{k},y_{r}^{k},t_{r}^{k};\boldsymbol{\theta}\right)\right)^{2}.\label{eq:loss_r3_NS}
\end{equation}

To conduct a sensitivity analysis on $M_{1}$ for the Navier--Stokes
equation, fifteen DG-PINN models are trained with the same $M_{1}$ values
as those specified in the sensitivity analysis on $M_{1}$ for the
heat equation. The accuracy of the predictions from the fifteen DG-PINN
models is evaluated by $\mathcal{R}_{t}$ shown in Fig. \ref{fig:NS_eq_stop}(a).
It can be seen that the values of $\mathcal{R}_{t}$ remain below
$2\times10^{-2}$ for $v\left(x,y,t\right)$ and below $6\times10^{-3}$
for $u\left(x,y,t\right)$. Besides, the APE between $\tilde{\beta_{1}}$
versus $\beta_{1}$ and $\tilde{\beta_{2}}$ versus $\beta_{2}$ for
each DG-PINN model is shown in Figure \ref{fig:NS_eq_stop}(b). It
is shown that the maximum APE between $\tilde{\beta_{1}}$ and $\beta_{1}$
is less than $0.3\%
$ for each $M_{1}$ and the maximum APE between $\tilde{\beta_{2}}$
and $\beta_{2}$ is less than $5\%$ for each $M_{1}$. In Figs. \ref{fig:NS_eq_stop}(c)
and (d), the observed pressure field $p\left(x,y,t\right)$ and predicted
pressure field $\hat{p}\left(x,y,t;\boldsymbol{\theta}\right)$ at
$t=1$ with mean centering are very similar. Mean centering is used because only the gradient of the pressure field drives the Navier-Stokes
equation, such that it remains unchanged with the addition of a constant
to the pressure field. It can be seen that the DG-PINN can solve the inverse problem of the Navier--Stokes equation accurately even with
a large range of $M_{1}$. These results emphasize the robustness
of the selection of $M_{1}$ for DG-PINNs.
\begin{center}
\begin{figure}[H]
\begin{centering}
\subfloat[]{\centering{}\includegraphics{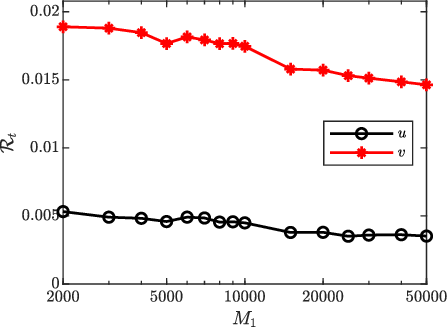}}\subfloat[]{\begin{centering}
\includegraphics{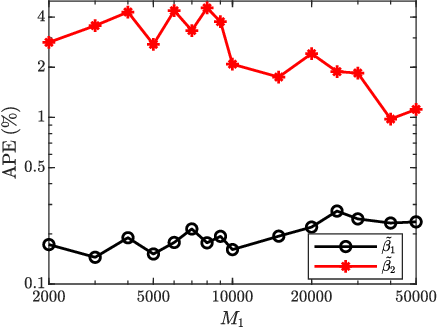}
\par\end{centering}
\centering{}}
\par\end{centering}
\centering{}\subfloat[]{\centering{}\includegraphics{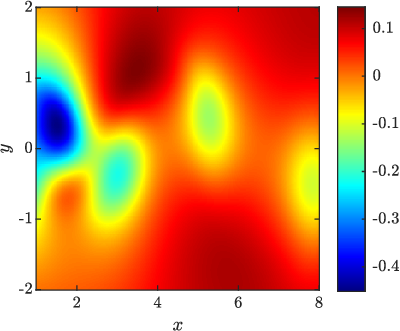}}\subfloat[]{\begin{centering}
\includegraphics{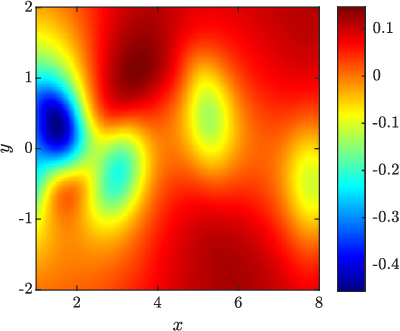}
\par\end{centering}
\centering{}}\caption{\label{fig:NS_eq_stop}Sensitivity analysis on $M_{1}$ for the Navier--Stokes
equation: (a) the relative $L^{2}$ errors between the testing data
and the corresponding predictions of the DG-PINN models, (b) the
absolute percentage error (APE) for $\tilde{\beta_{1}}$ versus $\beta_{1}$
and $\tilde{\beta_{2}}$ versus $\beta_{2}$, (c) the observed pressure
field $p\left(x,y,t\right)$ at $t=1$, and (d) the predicted pressure
field $\hat{p}\left(x,y,t;\boldsymbol{\theta}\right)$ at $t=1$.}
\end{figure}
\par\end{center}

To conduct a sensitivity analysis on $N_{d}$, fifteen DG-PINN models are trained using training datasets for $\mathcal{L}_{d1}\left(\boldsymbol{\theta}\right)$ and $\mathcal{L}_{d2}\left(\boldsymbol{\theta}\right)$ in the Navier--Stokes equation, with $N_{d}$ values matching those used in the sensitivity analysis for the heat equation. Figure. \ref{fig:NS_eq_data}(a) shows
that the values of $\mathcal{R}_{t}$ remain below $2\times10^{-2}$
for $v\left(x,y,t\right)$ and below $5\times10^{-3}$ for $u\left(x,y,t\right)$.
Besides, the APE between $\tilde{\beta_{1}}$ versus $\beta_{1}$
and $\tilde{\beta_{2}}$ versus $\beta_{2}$ for each DG-PINN model
is shown in Figure \ref{fig:NS_eq_data}(b). It is shown that the
maximum APE between $\tilde{\beta_{1}}$ and $\beta_{1}$ is less
than $0.3\%$ for each $N_{d}$ and the maximum APE between $\tilde{\beta_{2}}$
and $\beta_{2}$ is less than $2.6\%$ for each $N_{d}$. In Figs.
\ref{fig:NS_eq_data}(c) and (d), the observed pressure field $p\left(x,y,t\right)$
and predicted pressure field $\hat{p}\left(x,y,t;\boldsymbol{\theta}\right)$
at $t=1$ with mean centering are very similar. It can be seen that
DG-PINNs can solve the inverse problem of the Navier--Stokes equation
accurately even with the wide range of $N_{d}$. These results
highlights the robustness of the selection of $N_{d}$ for DG-PINNs.
\begin{center}
\begin{figure}[H]
\begin{centering}
\subfloat[]{\centering{}\includegraphics{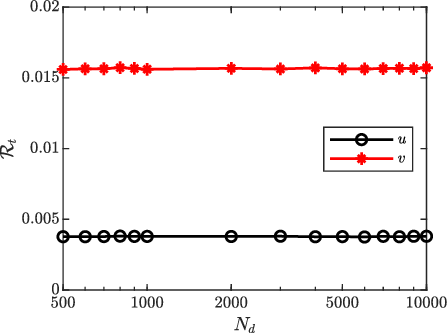}}\subfloat[]{\begin{centering}
\includegraphics{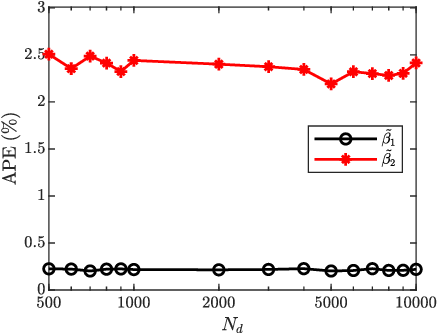}
\par\end{centering}
\centering{}}
\par\end{centering}
\centering{}\subfloat[]{\centering{}\includegraphics{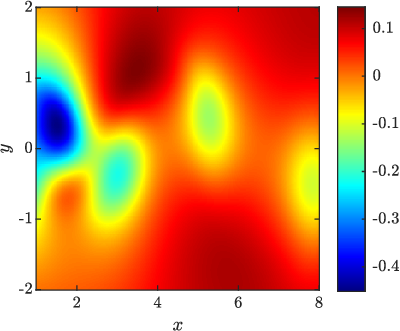}}\subfloat[]{\begin{centering}
\includegraphics{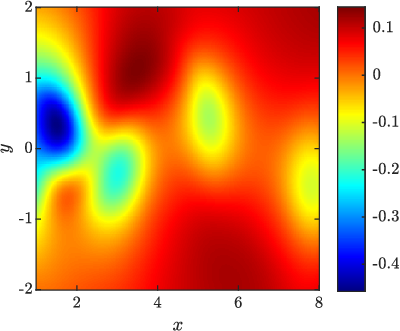}
\par\end{centering}
\centering{}}\caption{\label{fig:NS_eq_data}Sensitivity analysis on $N_{d}$ for the Navier--Stokes
equation: (a) the relative $L^{2}$ errors between the testing data
and the corresponding predictions of the DG-PINN models, (b) the absolute percentage error (APE) for $\tilde{\beta_{1}}$ versus $\beta_{1}$
and $\tilde{\beta_{2}}$ versus $\beta_{2}$, (c) the observed pressure
field $p\left(x,y,t\right)$ at $t=1$, and (d) the predicted pressure
field $\hat{p}\left(x,y,t;\boldsymbol{\theta}\right)$ at $t=1$.}
\end{figure}
\par\end{center}

To investigate the noise-robustness of DG-PINNs for the Navier--Stokes equation, two DG-PINN models are trained using observed data contaminated
with white Gaussian noise at SNRs of $40$ and $35$ dB, respectively. The relative $L^{2}$ errors $\mathcal{R}_{t}$ for $u\left(x,y,t\right)$
are calculated as $5.70\times10^{-3}$ and $1.06\times10^{-2}$ for SNRs of $40$ and $35$ dB, respectively. Similarly, for $v\left(x,y,t\right)$,
$\mathcal{R}_{t}$ is calculated as $2.14\times10^{-2}$ and $3.80\times10^{-2}$
for SNRs of $40$ and $35$ dB, respectively. Besides,
the APE between $\tilde{\beta_{1}}$ and $\beta_{1}$ is calculated
as $0.076\%$ and $0.22\%$ for SNRs of $40$ and $35$ dB,
respectively, and the APE between $\tilde{\beta_{2}}$ and $\beta_{2}$
is calculated as $0.136\%$ and $8.89\%$ for SNRs of $40$
and $35$ dB, respectively. These results indicate that DG-PINNs are
capable of solving the inverse problem of the Navier--Stokes equation
in the presence of noise contamination. However, the accuracy of $\beta_{2}$
estimation is more sensitive to noise, with a substantial decrease
in precision as the noise level increases.

To evaluate the efficiency of PINNs and DG-PINNs, both approaches
are applied in ten independent trials to solve the inverse problem
of the Navier--Stokes equation. The averaged $\mathcal{R}_{t}$, APE, and training time are listed in Table \ref{tab:NS_equation}. It is shown that
both $\mathcal{R}_{t}$ for $u\left(x,y,t\right)$ and $v\left(x,y,t\right)$
in DG-PINNs are lower than those in PINNs, indicating that DG-PINNs
provide more accurate predictions compared to PINNs. Regarding the APE for $\tilde{\beta_{1}}$
versus $\beta_{1}$ and $\tilde{\beta_{2}}$ versus $\beta_{2}$,
DG-PINNs exhibit a smaller error compared to PINNs, indicating that
DG-PINNs provide a more accurate estimation of $\beta_{1}$ and $\beta_{2}$.
Besides, the training time for DG-PINNs is significantly shorter than
that for PINNs. These results demonstrate that DG-PINNs are around nine
times faster than PINNs in solving the inverse problem of the Navier--Stokes
equation, achieving higher accuracy in both predictions and the estimation
of $\beta_{1}$ and $\beta_{2}$.
\begin{center}
\begin{table}[H]
\caption{\label{tab:NS_equation}Results of the comparative analysis between
DG-PINNs and PINNs for the Navier--Stokes equation.}

\centering{}%
\begin{tabular}{cccccc}
\toprule 
\multirow{2}{*}{Method} & \multicolumn{2}{c}{$\mathcal{R}_{t}$} & \multicolumn{2}{c}{APE} & \multirow{2}{*}{Training time}\tabularnewline
\cmidrule{2-5} \cmidrule{3-5} \cmidrule{4-5} \cmidrule{5-5} 
 & $u\left(x,y,t\right)$ & $v\left(x,y,t\right)$ & $\tilde{\beta_{1}}$ & $\tilde{\beta_{2}}$ & \tabularnewline
\midrule
PINN & $1.45\times10^{-2}$ & $6.17\times10^{-2}$ & $0.32\%$ & $13.83\%$ & $58.12$ min\tabularnewline
DG-PINN & $4.17\times10^{-3}$ & $1.66\times10^{-2}$ & $0.27\%$ & $2.29\%$ & $6.80$ min\tabularnewline
\bottomrule
\end{tabular}
\end{table}
\par\end{center}

\section{Concluding Remarks\label{sec:4}}

In this study, a novel framework for solving inverse problems in PDEs, referred to as the data-guided physics-informed neural networks
(DG-PINNs), is proposed. The DG-PINNs framework is structured two phases: a pre-training phase and a fine-tuning
phase. In the pre-training phase,
a neural network is trained to minimize the data loss. While in the
fine-tuning phase, the neural network is further optimized by integrating
fundamental physical laws into its loss function. The effectiveness,
noise-robustness, and efficiency of DG-PINNs in solving inverse problems
of PDEs are validated through a series of numerical investigations
on various PDEs, including the heat equation, wave equation, Euler--Bernoulli
beam equation and Navier--Stokes equation. For each equation, sensitivity
analyses of the DG-PINNs results are conducted focusing on two hyperparameters:
the maximum number of iterations in the pre-training phase and the
number of data points for the data loss. These sensitivity analyses
not only reveal the effectiveness of DG-PINNs in accurately solving
inverse problems in PDEs but also underscore the robustness of the
selection of these hyperparameters. Besides, investigations of noise-robustness
show that DG-PINNs are capable of accurately solving inverse problems
in PDEs in the presence of noise contamination. Furthermore, results
from the comparative analysis between DG-PINNs and PINNs show that
DG-PINNs offer a significant computational advantage over PINNs, demonstrating
faster computation times while maintaining high accuracy in the solution.
In future work, it would be valuable to explore the application of
DG-PINNs to a wider range of physical phenomena. It is worth noting that the two-phase framework can be extended to physics-informed
neural operators.

\section*{Acknowledgments}

The authors are grateful for the financial support from the National
Science Foundation through Grant No. CMMI-1762917.

\section*{Conflicts of Interest}

The authors declare no conflict of interest.

\bibliographystyle{elsarticle-num}
\bibliography{reference_zw}

\end{document}